\documentclass{article}
\usepackage{nips14submit_e}

\usepackage[hidelinks]{hyperref}
\usepackage{url}

\usepackage{graphicx}
\usepackage{caption, subcaption, capt-of}

\usepackage[round]{natbib}
\usepackage{cleveref}

\usepackage{algorithm}

\usepackage{listings}

\usepackage{amsmath,amsthm,amsfonts}

\usepackage{fancyvrb}
\usepackage{multicol}
\usepackage{comment}
\usepackage{mdframed}
\usepackage{algpseudocode}

\usepackage{cprotect}

\usepackage{authblk}

\usepackage{defs}

\title{Sublinear-Time Approximate MCMC Transitions for Probabilistic Programs}

\author[1]{Yutian Chen}
\author[2]{Vikash Mansinghka}
\author[1]{Zoubin Ghahramani}
\affil[1]{Machine Learning Group, Department of Engineering, University of Cambridge}
\affil[2]{Computer Science and Artificial Intelligence Laboratory, MIT}

\nipsfinalcopy 

\begin{document}

\maketitle

%

%

\begin{abstract}
Probabilistic programming languages can simplify the development of machine learning techniques, but only if inference is sufficiently scalable. Unfortunately, Bayesian parameter estimation for highly coupled models such as regressions and state-space models still scales poorly; each MCMC transition takes linear time in the number of observations. This paper describes a sublinear-time algorithm for making Metropolis-Hastings (MH) updates to latent variables in probabilistic programs. The approach generalizes recently introduced approximate MH techniques: instead of subsampling data items assumed to be independent, it subsamples edges in a dynamically constructed graphical model. It thus applies to a broader class of problems and interoperates with other general-purpose inference techniques. Empirical results, including confirmation of sublinear per-transition scaling, are presented for Bayesian logistic regression, nonlinear classification via joint Dirichlet process mixtures, and parameter estimation for stochastic volatility models (with state estimation via particle MCMC). All three applications use the same implementation, and each requires under 20 lines of probabilistic code.
\end{abstract}

\section{Introduction}

Machine learning methods can be difficult and time-consuming to design and implement. Probabilistic programming languages have the potential to mitigate these difficulties and make previously impractical approaches more manageable by formalizing and automating key aspects of modeling and inference. 
For example, the Stan platform can compactly represent hierarchical Bayesian models with complex structure and provide automatic HMC inference that has been successfully applied to problems with thousands of parameters \citep{stanmanual2014}. 
Similarly, the Picture language makes it possible to specify and solve challenging 3D vision problems \citep{picture}.
Other languages such as Church \citep{Goodman:2008}, BLOG \citep{blog2013}, Venture \citep{mansinghka2014venture} and Figaro \citep{Pfe09} are targeted at a broader class of modeling and inference problems.
All these languages go beyond standard toolkits for probabilistic graphical models in important ways.

Probabilistic programming thus provides an appealing setting in which to develop inference algorithms, but scalability is a significant challenge. Many probabilistic programming systems exhibit quadratic scaling on problems such as topic modeling \cite{BleiNgJordan03} and nonparametric mixture modeling \citep{Neal00}; for example, the Bher transformational compiler \citep{Goodman:2008} requires the entire model to be resimulated for each single-site MH transition. The asymptotic scaling of individual inference steps has been the focus of several research efforts. The Venture and BLOG languages overcome this by tracking dependencies between random choices, using the resulting factorization of the joint probability density over all variables to recover linear scaling in many common cases. The probabilistic execution trace (PET) graph that tracks dependencies in Venture also handles exchangeable coupling and thus supports algorithms based on $O(1)$ updates to sufficient statistics. Other approaches based on program analysis are also being explored \citep{yang2014generating}. The scaling problems are particularly severe in highly coupled parameter estimation problems, such as those arising in regression and state-space modeling. For these problems, there are no symmetries to exploit, so each MH update for a parameter requires $O(N)$ time.

This paper describes a sublinear-time algorithm for performing approximate MH transitions to latent variables in probabilistic programs with $O(N)$ outgoing dependencies. This algorithm generalizes ideas from recent work on approximate transition operators \citep{singh2012monte,korattikara2013austerity,bardenet2014towards}: instead of subsampling data items, it subsamples edges in a dynamically constructed graphical model, stochastically ignoring dependencies.  The proposed algorithm can be interleaved with state-of-the-art general-purpose inference algorithms for probabilistic programs and thus applies to problems with widely varying structures.

This paper contains three contributions. The first is the algorithm, including its integration into an inference programming language. The second is a new proof of ergodicity for the approximate Markov chain under milder conditions than \cite{korattikara2013austerity}, showing that the bias vanishes as the controlling parameter approaches 0. The third is an empirical demonstration of efficacy and broad applicability, via applications to parametric regression, nonparametric Bayesian mixtures of experts, and state-space models.

\section{Background on Inference in Probabilistic Programs}

Here we define the key terms relevant to the subsampled inference technique proposed in this paper. From the perspective of this paper, the function of the  background material is to exploit factorization in the joint density of program executions so that individual MH transitions do not have to traverse parts of the execution that have not changed. Due to space constraints, we cannot give a detailed or formal description of complete inference procedures for probabilistic programs. Readers are referred to \cite{mansinghka2014venture} for more details.

Probabilistic programming provides a convenient way for users to specify any probabilistic model with a potentially unbounded number of random variables in a formal language and conduct inference conditioned on observed data without implementing inference algorithms by themselves. A probabilistic program describes the generative process of the model with both deterministic or stochastic computations. A typical stochastic computation is to draw a random sample from some distribution. One execution of a program generates one realization of a probabilistic model. Specifically, the program returns a sequence of outputs in the order of execution, one from each computation, denoted by $x_1, x_2, \dots$. We use $\bx_{i:j}$ as a shorthand for $x_i,\dots,x_j$. The value of $x_i$ depends on its history $\bx_{1:i-1}$. So does the existence of one variable itself in the model if the computation appears in a control flow statement.

\begin{figure}[tb]
  \centering
\begin{Verbatim}[gobble=-30, numbers=left,numbersep=2pt, frame=single, xleftmargin=3mm,
xrightmargin = 3mm]
[assume b (bernoulli 0.5)]
[assume mu (if b 1 (gamma 1 1))]
[assume y (normal mu 0.1)]
[observe y 10.0]
\end{Verbatim}
\includegraphics[width=0.4\textwidth]{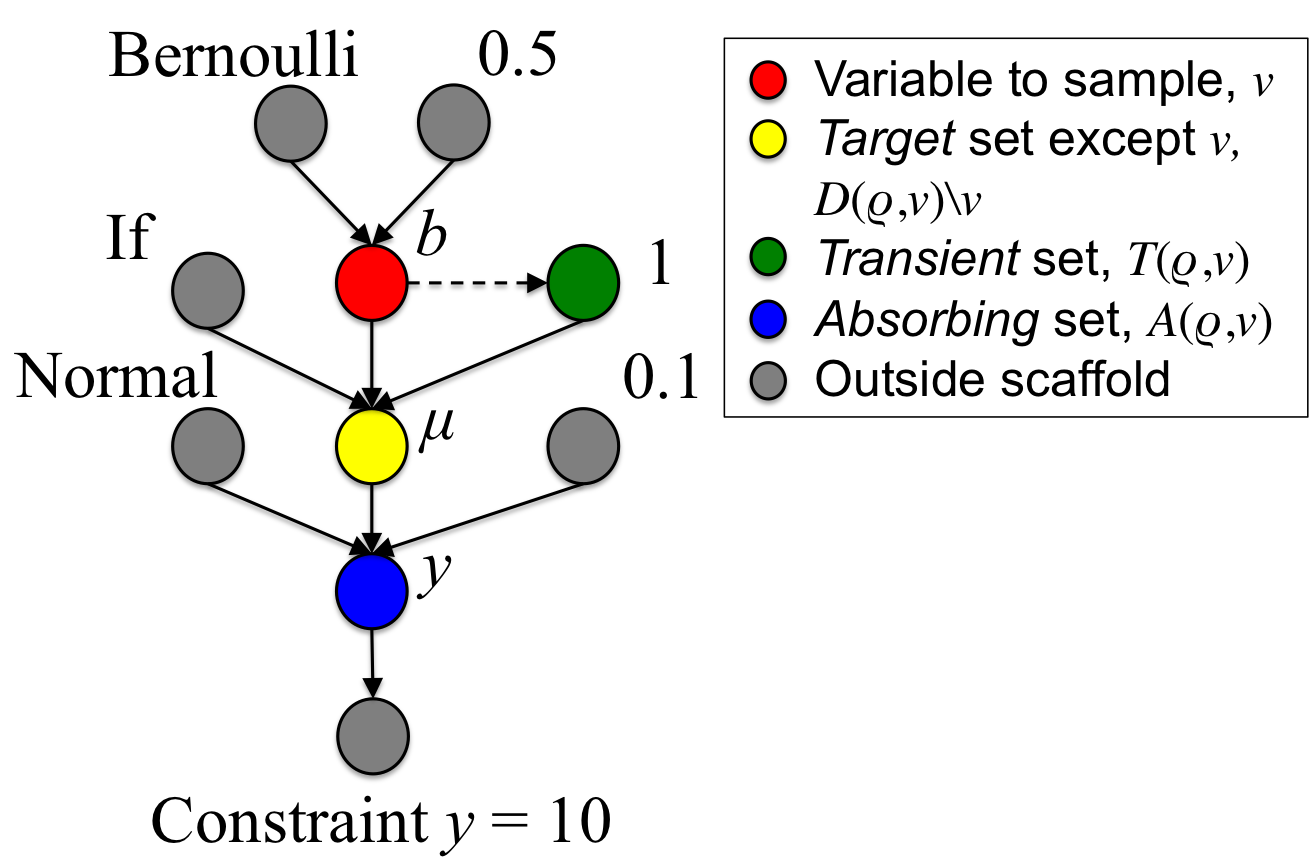}
\cprotect\caption{Example of a probabilistic program and its PET when $b=True$. Solid and dashed arrows represent edges in $E_\mathrm{s}$ and $E_\mathrm{e}$ respectively. Colored nodes specify the scaffold for sampling $b$ which is defined later in Sec.~\ref{sec:mh_pet}.}
\label{fig:pet_example}
\end{figure}

An example of a probabilistic program in a Lisp style language syntax is shown in the top of Fig.~\ref{fig:pet_example}. \verb|assume| statements specify the generative model and \verb|observe| statements specify observations, i.e.\ constraints on random variables. The program first draws a Bernoulli random variable $b$, assigns the value of variable $\mu$ with 1 if $b=True$ or a Gamma random variable otherwise. It then draws a normal random variable $y$ with mean equal to $\mu$ and specifies an observation $10$. \verb|bernoulli|, \verb|gamma| and \verb|normal| are stochastic computations, and \verb|if| is a deterministic control flow computation. The execution of a program can be represented as a directed graph defined as follows:

\textbf{Definition 1}. A {\em probabilistic execution trace} (PET or trace) is a directed graph, $\rho := (V, E_{\text{e}} \cup E_{\text{s}})$, representing a single run of a probabilistic program, where the set of nodes $V := \{i\}$ represents all executed computations, $E_{\text{e}} := \{(i, j): x_i$ is the last variable in $\bx_{1:j-1}$ whose value determines the existence of $x_j\}$ represents all {\em existential} dependencies, and $E_{\text{s}} := \{(x_i, x_j): i<j, x_i \not\!\perp\!\!\!\perp x_j | \bx_{1:j-1\backslash i}\}$ represents all {\em statistical} dependencies between the value of two nodes.

$E_{\text{s}}$ includes both deterministic and stochastic dependencies while $E_{\text{e}}$ determines the existence of nodes in $V$. Fig.~\ref{fig:pet_example} shows a trace where $b=True$. Because the \verb|if| statement chooses the first branch, the computation of sampling a Gamma random variable, \verb|(gamma 1 1)|, is not executed and therefore does not exist in the trace.


Given $V$, which depends on $E_{\text{e}}$, and their statistical dependencies $E_{\text{s}}$, the probability of generating the trace $\rho$ has the same factorization as a Bayesian network of $(V, E_{\text{s}})$:
\begin{equation}
p(\rho) = \prod_{n \in V} p(x_n | \mathrm{Par}_{\rho}(n)) \label{eq:p_pet}
\end{equation}
where $\mathrm{Par}_{\rho}(n) := \{x_{n'} : n' \in V, (n', n) \in E_{\text{s}}(\rho)\}$ is the parent set of node $n$ in trace $\rho$.

\subsection{Metropolis-Hastings Sampling Algorithm on PETs}\label{sec:mh_pet}

Now we describe how to do inference for a probabilistic program with a single-site MH sampling algorithm and its computational complexity. Denote the variable to sample by $x_v, v \in V$. After proposing a new value for $x_v$, to maintain consistency of the trace we have to propagate the change throughout the descendants of $v$ whose value or existence depends on $x_v$ deterministically.

\textbf{Definition 2}. The {\em target} set, $D(\rho,v)$, is the set of $v$ and all of its descendants whose computation is always executed for any value of $x_v$ but their value depends on $x_v$ deterministically.

\textbf{Definition 3}. The {\em transient} set, $T(\rho,v)$, is the set of all the variables whose existence depends on the value of $D(\rho,v)$, $T(\rho,v):=\{n\in V: \exists n'\in D(v,\rho), s.t.\ (n',n)\in E_e\}$.

The change of $x_v$ can be absorbed by children of $D$ or $T$ that represent random computations in the sense that we do not need to propose a new value for those nodes (e.g.\ node $y$ in Fig.~\ref{fig:pet_example}) to maintain the consistency of a trace. However, to evaluate the probability of the new trace, we still need to recompute the probability of generating those nodes with the new value of $x_v$.

\textbf{Definition 4}. The {\em absorbing} set is
$A(\rho,v):=\{n\in V\backslash(D \cup T): \exists n'\in D \cup T, s.t.\ n'\in \text{Par}_{\rho}(n)\}$.

For a probabilistic program that describes a regular Bayesian network model without existential and deterministic dependencies, we have the following relationships:
\begin{equation}
D(\rho,v)=\{v\},~~ T(\rho,v) = \emptyset,~~ A(\rho,v) = \text{Children}(v) \nn
\end{equation}

Lastly, a scaffold is defined as the union of all the three sets:

\textbf{Definition 5}. The {\em scaffold} of $v$ in trace $\rho$ is the set of nodes: $s(\rho,v):=D(\rho,v) \cup T(\rho,v) \cup A(\rho,v)$.

One may observe that the scaffold is the set of all the variables whose conditional distribution in Eq.~\ref{eq:p_pet} may change with a new proposal for $x_v$. Particularly when the value of $v$ does not affect the structure of $\rho$, i.e.\ $T(\rho,v)=\emptyset$, the union of $s(\rho,v)$ and its parent set excluding $D$, $(\cup_{n\in s}\{\text{Par}_{\rho}(n)\} \cup s) \backslash D$, is equivalent to the Markov blanket of $v$ in the corresponding Bayesian network $(V, E_{\text{s}})$.

Now we have defined all the ingredients to describe the MH algorithm on a PET. Alg.~\ref{alg:mh} describes the steps to sample variable $x_v$ given a proposal distribution $q$. A new trace is proposed in step \ref{step:construct_s}-\ref{step:regen} and accepted/rejected in the remaining steps. The proposal distribution can be the prior distribution from the program or a customized distribution provided by a user.
\begin{algorithm}
\caption{Metropolis-Hastings Algorithm with Scaffolds}\label{alg:mh}
\begin{algorithmic}[1]
\Procedure{MH}{$v$, $\rho$, $q$}
  \State Construct $s(\rho, v)$\label{step:construct_s}.
  \State $detach$: removes $D\cup T$ from $\rho$\label{step:detach}.
  \State $regenerate$: proposes new values for $D$ with $q$ and generates a new transient set, $T'$, depending on the new values\label{step:regen}.
  \State Compute acceptance probability $P_a$\label{step:P_a}.
  \State Sample $u\sim \mathrm{Uniform}[0,1]$.
  \If {$u \leq P_a$}
    \State Accept the new trace.
  \Else
    \State $detach$ $D \cup T'$ and restore old values of $D\cup T$.
  \EndIf
\EndProcedure
\end{algorithmic}
\end{algorithm}

Denote the proposed trace with $\rho'$ and the new scaffold with $s'$. As the structure of $\rho$ does not change except for the transient set, we get $s' = D \cup T' \cup A$. The reverse proposal of MH includes \textit{detaching} $D \cup T'$ from $\rho'$, and \textit{regenerating} old values of $D$ and $T$. The acceptance probability in Step \ref{step:P_a} is then computed as
\begin{align}
P_a(\rho, \rho') &= \min\left\{1, \frac{p(\rho') \prod_{n \in D \cup T} q(x_n|\rho)}{p(\rho) \prod_{n \in D \cup T'} q(x'_n|\rho')} \right\} \\
&= \min\left\{1, \prod_{n\in D} \frac{p(x'_n|\mathrm{Par}_{\rho'}(n))}{q(x'_n|\rho')} \frac{q(x_n|\rho)}{p(x_n|\mathrm{Par}_{\rho}(n))} \right.\nn\\
& \prod_{n\in A} \frac{p(x_n|\mathrm{Par}_{\rho'}(n))}{p(x_n|\mathrm{Par}_{\rho}(n))} \prod_{n\in T} \frac{q(x_n|\rho)}{p(x_n|\mathrm{Par}_{\rho}(n))}\nn\\
& \left. \prod_{n\in T'} \frac{p(x'_n|\mathrm{Par}_{\rho'}(n))}{q(x'_n|\rho')}\right\} \label{eq:P_a}
\end{align}
where we plugged in Eq.~\ref{eq:p_pet} in the second line. $q(x'_n|\rho')$ denotes the proposal distribution of node $n$ given all the nodes in the new trace $\rho'$ that have been generated before $n$. Any factors for nodes outside $s \cup s'(= D \cup A \cup T \cup T')$ are canceled, consistent with our previous observation about the scaffold. The factors in Eq.~\ref{eq:P_a} depending on $\rho'$ are computed on the fly during Step \ref{step:regen} and the factors depending on $\rho$ are computed during Step \ref{step:detach}. For any node $n \in s \cup s'$, we denote by $w_n$ the product of all the factors in Eq.~\ref{eq:P_a} indexed by $n$. Then we can simplify the acceptance probability as
\begin{equation}
P_a(\rho, \rho') = \min \left\{1, \prod_{n\in s \cup s'} w_n\right\} \label{eq:P_a_w}
\end{equation}

\subsection{Computational Complexity of MH Transitions}

Eq.~\ref{eq:P_a_w} makes it clear that the running time of one MH transition scales linearly with the size of the scaffold. For a \textit{local} random variable with a constant size of the scaffold/Markov blanket, it takes $O(1)$ time to run the MH transition. However, for a \textit{global} random variable, e.g.\ a parameter coupled with all the observations, the size of the scaffold scales with the size of the entire data set. Therefore, in a model with $D$ global variables and $N$ observations, it takes  $O(DN)$ time to run MH for just one sweep of all the global variables. That could be unacceptably slow for a problem with a large data set. Given a limited budget on the computational time, a slow transition operator leads to a small number of samples and therefore a large variance in the Monte Carlo estimate to any quantities of interest.

\begin{figure*}[tb!]
  \centering
  \begin{subfigure}[t]{0.3\textwidth}
   \centering
   \includegraphics[width=\textwidth]{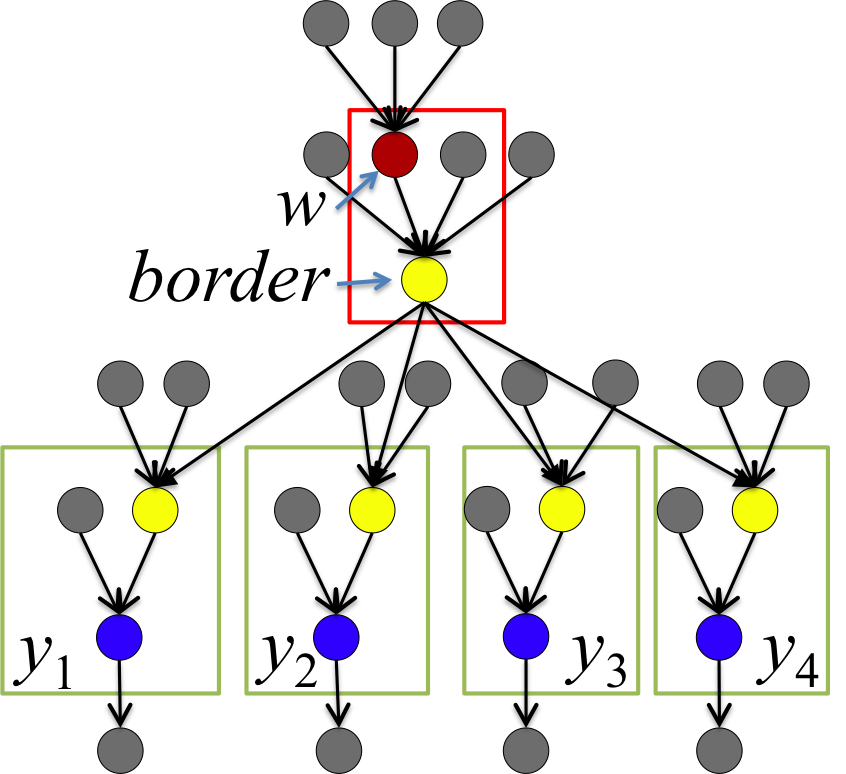}
   \caption{} \label{fig:scaffold_bayeslr_ppt}
 \end{subfigure}
  \begin{subfigure}[t]{0.22\textwidth}
    \centering
    \includegraphics[width=\textwidth]{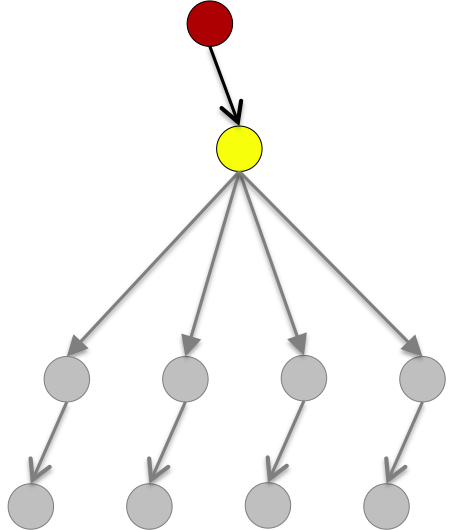}
    \caption{}
    \label{fig:scaffold_2}
  \end{subfigure}
  \begin{subfigure}[t]{0.22\textwidth}
    \centering
    \includegraphics[width=\textwidth]{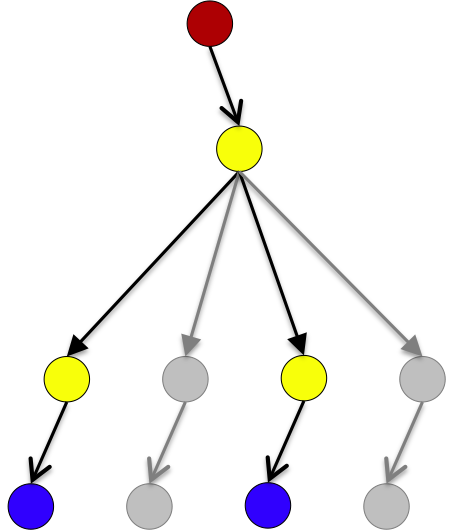}
    \caption{}
    \label{fig:scaffold_3}
  \end{subfigure}
  \begin{subfigure}[t]{0.22\textwidth}
    \centering
    \includegraphics[width=\textwidth]{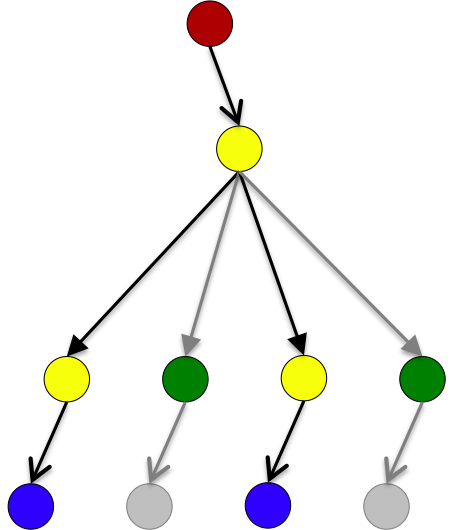}
    \caption{}
    \label{fig:scaffold_4}
  \end{subfigure}
  \caption{(a) The PET of a Bayesian logistic regression model with 4 observations, colored to encode the scaffold for the weights $\bw$; the probabilistic program is shown in Fig.~\ref{fig:venture_bayeslr}. The global section is notated with a red box and local sections with green boxes. (b, c) Subsampled scaffold with no and two local sections. (d) Nodes whose values are stale after a subsampled transition. See Sec.~\ref{sec:stale_nodes} for details.}
  \label{fig:subscaffold}
  \vspace{-.3cm}
\end{figure*}

\section{Sublinear Time Inference via Subsampling Scaffolds}
This paper presents an approximate MH transition for a global variable whose runtime scales sublinearly with the size of data, $o(N)$. We show that the approximate Markov chain is still uniformly ergodic under mild conditions and the induced bias of the stationary distributed is controllable. As we can now generate more samples given the same computational budget, it is possible to improve the accuracy of the MCMC estimate by trading off some bias for a much smaller variance, as confirmed in the experimental section.

\subsection{A Global/Local Partition of the Scaffold}\label{sec:partition}

We start with an observation of the scaffold for global variables. Fig.~\ref{fig:scaffold_bayeslr_ppt} shows an example of the scaffold for sampling the weights of a Bayesian logistic regression model with the program specified in the first 7 lines of Fig.~\ref{fig:venture_bayeslr}. The scaffold $s(\rho,v)$ can be partitioned into two parts:
\begin{enumerate}
\item A global section, denoted as $global$, that contains $v$ and dependencies that do not scale with $N$.
\item A set of $N$ local sections, $\{local_n\}_{n=1}^N$, which share a similar structure and represent the $N$ dependencies.
\end{enumerate}
The size of the scaffold grows with the number of local sections while the size of each section remains constant. 

Our algorithm applies to scaffolds of this structure, where in addition we assume all the local dependencies are connected from $v$ through a common node with a single link, which is $v$ itself or other node in $D(\rho, v)$. Under this assumptions, the common node, e.g. the top yellow node in Fig.~\ref{fig:scaffold_2}, is the border that separates {\em global} from {\em local} sections and has $N$ children.
We also assume that $T(\rho, v) = \emptyset$, i.e.\ making proposals for $x_v$ does not change the structure of the PET. Note that this does {\em not} exclude models with dynamically evolving structure; only approximate transitions are prohibited from introducing structural changes. These are mild restrictions, still admitting a broad class of probabilistic programs.

Formally, we provide constructive definitions for the terms above

\textbf{Definition 6}. The {\em border} node, $b(s, v)$, of a scaffold $s(\rho, v)$, is the first descendant of $v$ with multiple branches in $s$.

\textbf{Definition 7}. The {\em global} section of a scaffold $s(\rho, v)$ is: $global(\rho, v) := s(\rho, v) \backslash desendants(\rho, b)$.

\textbf{Definition 8} The {\em local} section of a scaffold $s(\rho, v)$ associated with $i$'th child of $b$, $c_i$, is: $local_i := s \cap (c_i \cup desendants(\rho, c_i))$.

The scaffold is then partitioned into one {\em global} and $N$ {\em local} sections mutually exclusively.

\subsection{Austerity MCMC for Probabilistic Programs with Partitioned Scaffolds}

\cite{korattikara2013austerity} proposed the austerity idea, a sublinear-time algorithm to incrementally approximate the acceptance probability until the approximation error is under a tolerance level. They applied the algorithm to the posterior inference problem for Bayesian models with $N$ \emph{iid} observations. We adopt this idea for the inference problem on PETs by proposing a sublinear-time algorithm for approximate MH transitions on partitioned scaffolds.

Given a partition of the scaffold, we can factorize the acceptance probability in Eq.~\ref{eq:P_a_w} as
\begin{equation}
P_a = \min\left[1, \left(\prod_{n\in global} w_n\right) \prod_{i=1}^N \left( \prod_{n\in local_i}w_n \right)\right] \label{eq:P_a_partition}
\end{equation}
After drawing a uniform random variable $u$, the inequality $u < P_a$ is equivalent to $\mu_0 < \mu$ with $\mu_0$ and $\mu$ defined as follows:
\begin{equation}
\mu_0 = \frac{1}{N} ( \log u - \sum_{n\in global} \log w_n),
\mu = \frac{1}{N} \sum_{i=1}^N l_i, \label{eq:mu}
\end{equation}
where $l_i = \sum_{n\in local_i} \log w_n$.
The term $\mu_0$ is usually cheap to compute, while $\mu$ is an expensive average over $N$ terms when $N$ is large. We follow \cite{korattikara2013austerity} and reformulate the decision problem as a statistical test with the hypothesis $H_1: \mu > \mu_0$, and $H_2: \mu < \mu_0$. Given $\mu_0$, the set $\{l_i\}$ and a tolerance level $\eps$, we conduct the hypothesis testing by iteratively sampling mini-batches of $m$ local sections, and updating the estimate of $\mu$ until a high confident result is reached. The algorithm is given in Alg.~\ref{alg:submh_mu}. The last term in $\sqrt.$ in step 7 is called the finite population correction factor introduced by sampling without replacement. In step \ref{step:t-test}, if $s_l=0$, we do not run the $t$-test but continue to draw another subset until $n=N$. That prevents a false decision in early stages when a small subset of $\{l_i\}$ includes all equal numbers by chance. For problems with a large data set, the approximate MH method requires only a small fraction of the data to make a high quality decision, and therefore can significantly speed up the MH algorithm.

\begin{algorithm}
\caption{Sequential Test for MH Decison}\label{alg:submh_mu}
\begin{algorithmic}[1]
\Procedure{SequentialTest}{$\mu_0$, set $\cX=\{l_i\}$ of a size $N$, mini-batch size $m$, error tolerance $\eps$}
  \State Set of observed data $\cY\gets \emptyset$, $n\gets 0$.
  \Repeat
    	\State Draw a subset of $m$ $l_i$'s, $X_m \subseteq \cX$
    	\State $\cX \gets \cX \backslash X_m$, $\cY \gets \cY \cup X_m$, $n\gets n+m$
    \State Estimate mean, $\hat{\mu}$, and std, $s_l$, of $\cY$.
    \State std of $\hat{\mu}$ is $s\gets \frac{s_l}{\sqrt{n}}\sqrt{1-(n-1)/(N-1)}$ 
  \Until{p-value of $t_{n-1}(|\frac{\hat{\mu}-\mu_0}{s}|)$ is lower than $\eps$}\label{step:t-test}
  \State \textbf{Accept} $H_1$ \textbf{if} $\hat{\mu}>\mu_0$ \textbf{else} $H_2$
\EndProcedure
\end{algorithmic}
\end{algorithm}


By comparing the definition of $\mu_0$ and $\mu$ with those in \cite{korattikara2013austerity}, one may find out that $\mu_0$ in both methods contains the same prior and proposal distribution for $x_v$. However, the quantity $\mu$ in our method has a more general meaning. In \cite{korattikara2013austerity}, each $l_i$ refers to the log-ratio of the likelihood of one \emph{iid} data point, while in our PET setting each $l_i$ is the product of $w$'s associated with a local partition. They do not have to be observed data, and can even have strong dependence between each other; our third experiment illustrates this case.

\textbf{Remark}. The same error analysis in \cite{korattikara2013austerity} applies to our algorithm for PETs even at the presence of dependencies between local sections in the model distribution.

That remark follows the fact that the sequential test procedure in Alg.~\ref{alg:submh_mu} exploits the statistical property of the randomly subsampled $l_i$ when the entire set $\{l_i\}$ is given. \textit{Conditioned} on $\{l_i\}$, the samples randomly drawn without replacement are independent with each other except for the correlation introduced by the ``without replacement" part. Therefore, error induced by the sequential test does not depend on whether there exist dependencies among $l_i$'s in the prior distribution defined by the probabilistic program.

\subsection{Robustness for General Problems}

It is worth noting that the error of the sequential test depends on the \textit{marginal} distribution of $\{l_i\}$'s. This is because our algorithm is based on student-$t$ test. When there exist many outliers with extremely large values, as demonstrated by a synthetic counter-example in \cite{bardenet2014towards}, the central limit theorem breaks down on the subset of $\{l_i\}$. In that case, our approximate MH transition might lead to noticeable bias in the stationary distribution. Despite the potential failure in the synthetic example, the austerity MH algorithm has shown robust performance in all the real-data experiments from both \cite{korattikara2013austerity} and this paper with a proper size of a mini-batch. 

Our software can provide a normality test for the distribution of the estimated mean $\hat\mu$ in trial runs and produce an auto-generated comparison between the performance of the approximate MH and regular inference. One advantage of a probabilistic programming implementation is that users do not need to implement safeguards themselves. In practice, we suggest applying our proposed algorithm on variables with a single type of dependence to reduce the risk of outliers.

Additionally, we prove that even when the condition for the central limit theorem does not hold, the bias of the approximate Markov chain still diminishes as the tolerance parameter $\eps$ in Alg.~\ref{alg:submh_mu} approaches 0 under mild conditions.

Let $\ta$ and $\ta^*$ be the current and proposed value of $x_v$, $m$ be the size of a mini-batch, and $P_{a,\eps}(\ta, \ta^*)$ be the acceptance probability using the sequential test with $\eps$. We have the following theorem with proof in the supplementary.
\begin{theorem}\label{thm:max_error}
If either (1) the domain of $\ta$, $\Theta$, is compact and the likelihood function is continuous w.r.t.\ $\ta$, or (2) $\Theta$, is a finite set, then there exists a function $\de(\eps)$ such that $|P_{a,\eps}(\ta, \ta^*) - P_a(\ta, \ta^*)| \leq \de(\eps), \forall \ta, \ta^* \in \Theta$, and $\de(\eps) \rightarrow 0$ as $\eps \rightarrow 0$.
\end{theorem}

This theorem shows a uniform convergence of the approximate MH transition to the exact transition. The compactness condition might be violated if $\ta$ has an unbounded domain. However, its impact to the bias of the stationary distribution could be negligible when the actual samples of $\ta$ reside in a finite region for any sufficiently small $\eps$. Alternatively, one can use a truncated prior distribution for $\ta$.

We can further prove the uniform ergodicity of our approximate Markov chain as follows.
\begin{corollary}\label{cor:ergodicity}
If either condition in Thm.~\ref{thm:max_error} holds and the exact Markov chain satisfies the regularity assumptions 1-3 in Sec.~3.2 of \cite{pillai2014ergodicity}, the Markov chain of the approximate MH algorithm is uniformly ergodic for any sufficiently small $\eps$. Its stationary distribution approaches the target distribution as $\eps\rightarrow 0$.
\end{corollary}

\subsection{Implementation with Scaffold Subsampling}

In order to obtain sublinear running time, we should avoid any operations with a $O(N)$ complexity, including the construction of the entire scaffold. Our proposed algorithm interleaves the scaffold construction with the {\em detach} \& {\em regenerate} operations. This way the local sections of the scaffold will not be created until the sequential test requires more data to improve its estimate. Let Detach\&Regen($A$) be the procedure to detach and regenerate a set of nodes $A\subseteq s$ and return $\sum_{n\in A}\log w_n$. Our subsampled MH algorithm is given in Alg.~\ref{alg:submh}.
Fig.~\ref{fig:subscaffold}(\subref{fig:scaffold_2}, \subref{fig:scaffold_3}) illustrate the state after step \ref{alg:global_partition} and after two local partitions are constructed respectively.
\begin{algorithm}
\caption{Sublinear-Time Metropolis-Hastings Algorithm with Scaffolds}\label{alg:submh}
\begin{algorithmic}[1]
\Procedure{SubsampledMH}{$v$, $\rho$, $q$, $m$, $\eps$}
  \State Sample $u\sim \mathrm{Uniform}[0,1]$.
  \State Find $b(s, v)$ as Def.~6, $N \gets$ \#($b$'s children)
  \State Construct $global$ as Def.~7 by contructing a regular $s(\rho,v)$ without reaching beyond $b$\label{alg:global_partition}
  \State $\sum_{n\in global}\log w_n \gets \mathrm{Detach\&Regen}(global)$
  \State Compute $\mu_0$ with Eq.~\ref{eq:mu}, $\cY\gets \emptyset$, $n\gets 0$
  \Repeat
    \State Sample $m$ children of $b$'s w/o replacement, $\{c_i\}$.
    \ForAll{$c_i$}
      \State Construct $local_i$ as Def.~8 by resuming the construction of $s(\rho,v)$ from $c_i$
      \State $l_i \gets \mathrm{Detach\&Regen}(local_i)$
    \EndFor
    \State $\cY\gets \cY\cup \{l_i\}$, update $n$, $\hat{\mu}$, and $s$ as in Alg.~\ref{alg:submh_mu}
  \Until{The p-value falls below $\eps$ as in Alg.~\ref{alg:submh_mu}}
  \If {$\hat{\mu}>\mu_0$}
    \State Accept the new trace.
  \Else
    \State $detach$ and restore old values of $global$.
  \EndIf
\EndProcedure
\end{algorithmic}
\end{algorithm}

\subsection{Updating stale nodes on demand} \label{sec:stale_nodes}
When a proposed move is accepted, all the deterministic nodes, $D(\rho, v)$, should be updated accordingly in the standard MH algorithm. However, with the subsampling approach it is not guaranteed to update all \textit{local} sections if some of them have not been constructed. This may leave some nodes with old values and break the deterministic dependencies. The green nodes in Fig.~\ref{fig:scaffold_4} are examples of the stale nodes. In order to solve the broken consistency while retaining sublinear runtime, we proposed a lazy updating approach: instead of updating deterministic nodes in local sections immediately after a proposal is accepted/reject, we leave the stale deterministic nodes as is but whenever it is to be accessed later, we first update their values and then process the node.


\section{Experiments}\label{sec:experiments}
We implemented our algorithm in a lightweight research variant of the Venture probabilistic programming system, written in unoptimized Python, and applied it to multiple problems. Asymptotic scaling and relative comparisons between standard and subsampled MH are meaningful, but absolute runtimes reflect large constant factor overheads that are straightforward to reduce.
\begin{table}[tbh]
\centering
\caption{Overview of models used in experiments, and scaling parameters for exact MH. $N_k:=\sum_{i=1}^N \eI[z_i=k]$.}
\label{tab:post}
\begin{tabular}{c|c|c}
Model & Domain of Sublinear MH & Scaling \\
\hline
BayesLR & $\bw \sim p(\bw)\prod_{i=1}^N \mathrm{Logit}(y|\bx_i, \bw)$ & $N$ \\
JointDPM & $\bw_k \sim p(\bw_k)\prod_{i:z_i=k} \mathrm{Logit}(y|\bx_i, \bw)$ & $N_k$ \\
SV & $\phi/\sg \sim p(\phi/\sg)\prod_{t=1}^T\cN(h_t|\phi h_{t-1},\sg^2)$ & $T$ \\
\end{tabular}
\end{table}
\subsection{Bayesian Logistic Regression}
\begin{figure}[tb]
\centering
 \includegraphics[width=.4\textwidth]{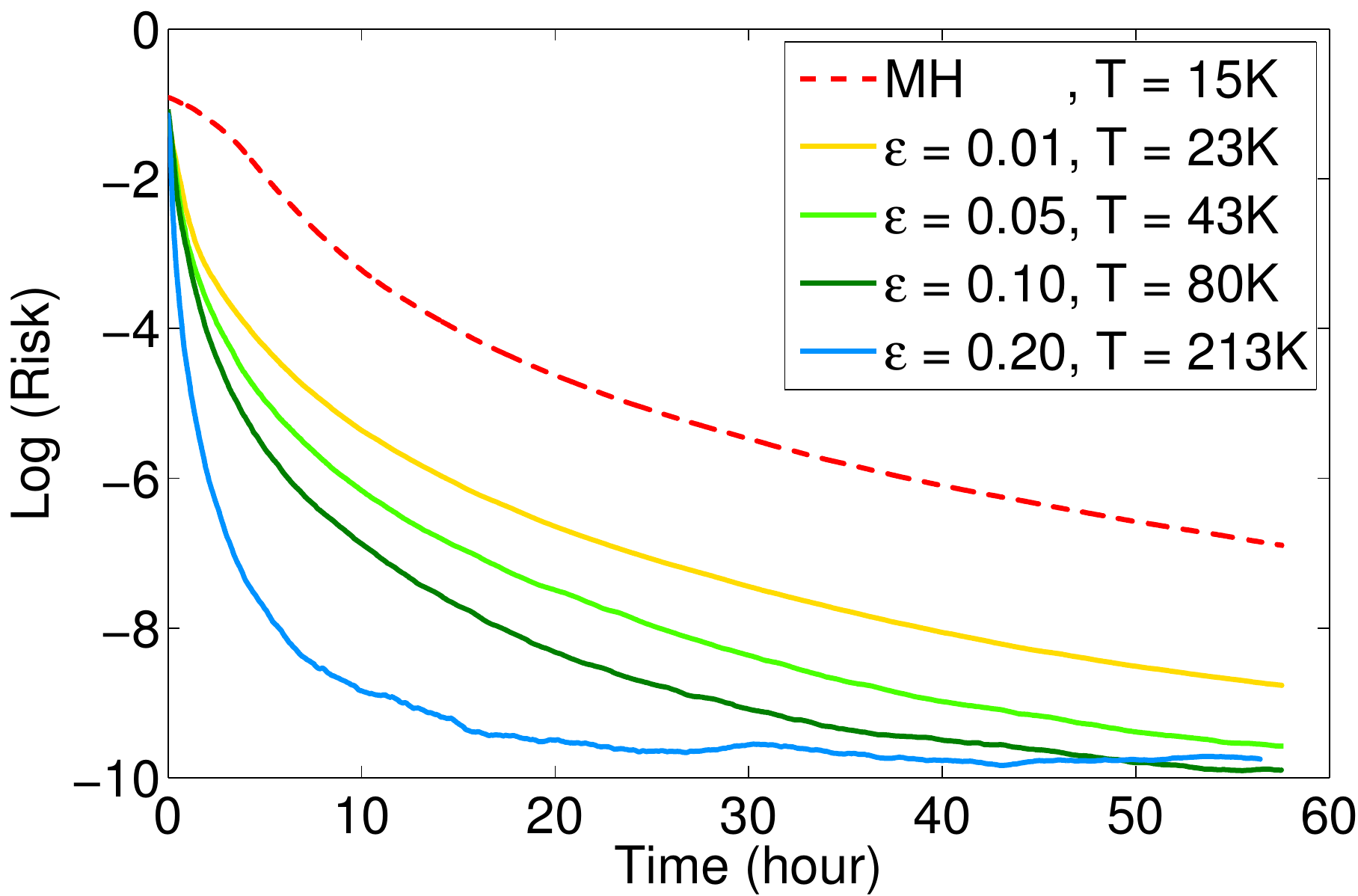}
 \caption{Risk of predictive mean vs.~running time for Bayesian logistic regression on MNIST. T in the legend denotes the number of samples drawn in about 60 hours.}
 \label{fig:mnist_pred}
 \vspace{-0.3cm}
\end{figure}

We first demonstrate that our general-purpose implementation recovers the time/accuracy advantages of the custom implementation of austery for Bayesian logistic regression from \cite{korattikara2013austerity}. Our Bayesian logistic regression model uses a standard isotropic Gaussian prior on the weights:
\begin{equation}
\bw \sim \cN(\bzero, 0.1\bI_D), \quad 
y_i \overset{\textnormal{iid}}\sim \mathrm{Logit}(y|\bx_i, \bw)
\end{equation}
We evaluate performance on a classification task on the MNIST digit image data. We train on 12214 images of `7' and `9', each transformed to a 50-dimensional feature vector via normalization and principal component analysis. We evaluate predictive accuracy versus computation time on 2037 test images. The Venture programs for the model and inference is given in Fig.~\ref{fig:venture_bayeslr}. We use the same parameter and data settings as in \cite{korattikara2013austerity}, and run the inference algorithm with the same random walk proposal distribution. We use a smaller mini-batch size of 100. The risk of predictive mean over an extensive run over 50 hours are given in Fig.~\ref{fig:mnist_pred}. See \cite{korattikara2013austerity} for the definition of risk. The predictive performance of subsampled MH increases significantly faster than the standard MH. It can make more than one order of magnitude more transitions, and takes 5 hours to reach the risk achieved by standard MH after 50 hours.
\begin{figure}
\vspace{-0.1in}
\small\begin{Verbatim}[gobble=-30, numbers=left,numbersep=2pt, frame=single, xleftmargin=3mm,
xrightmargin = 3mm]
[assume w (scope_include 'w 0
           (multivariate_normal {mu} {Sig}))]             
[assume y_x (lambda (x) (bernoulli (linear_logistic w x)))]

for n in 1...N: #load data
  # features X[n] = [x0_n x1_n ...], class Y[n]
  [observe (y_x {X[n]}) {Y[n]}]

# do T iterations of subsampled MH with a Gaussian drift
# proposal of bandwidth sig
[infer (subsampled_mh w all {nbatch} {eps} 'drift {sig} {T}]
[infer (mh w all {T})] #standard MH provided for reference
\end{Verbatim}
\cprotect\caption{Probabilistic program for the Bayesian logistic regression model, data, and subsampled inference scheme. $\{\}$ denotes external parameters. (\verb|scope_include| 'w 0 x) is a deterministic computation that adds label $w$ to $x$ for the \verb|infer| statement to specify which variable to sample.}
\label{fig:venture_bayeslr}
\end{figure}

\subsubsection{Sublinearity}
\begin{figure*}[tb]
\centering
\begin{subfigure}[b]{0.32\textwidth}
  \includegraphics[width=\textwidth]{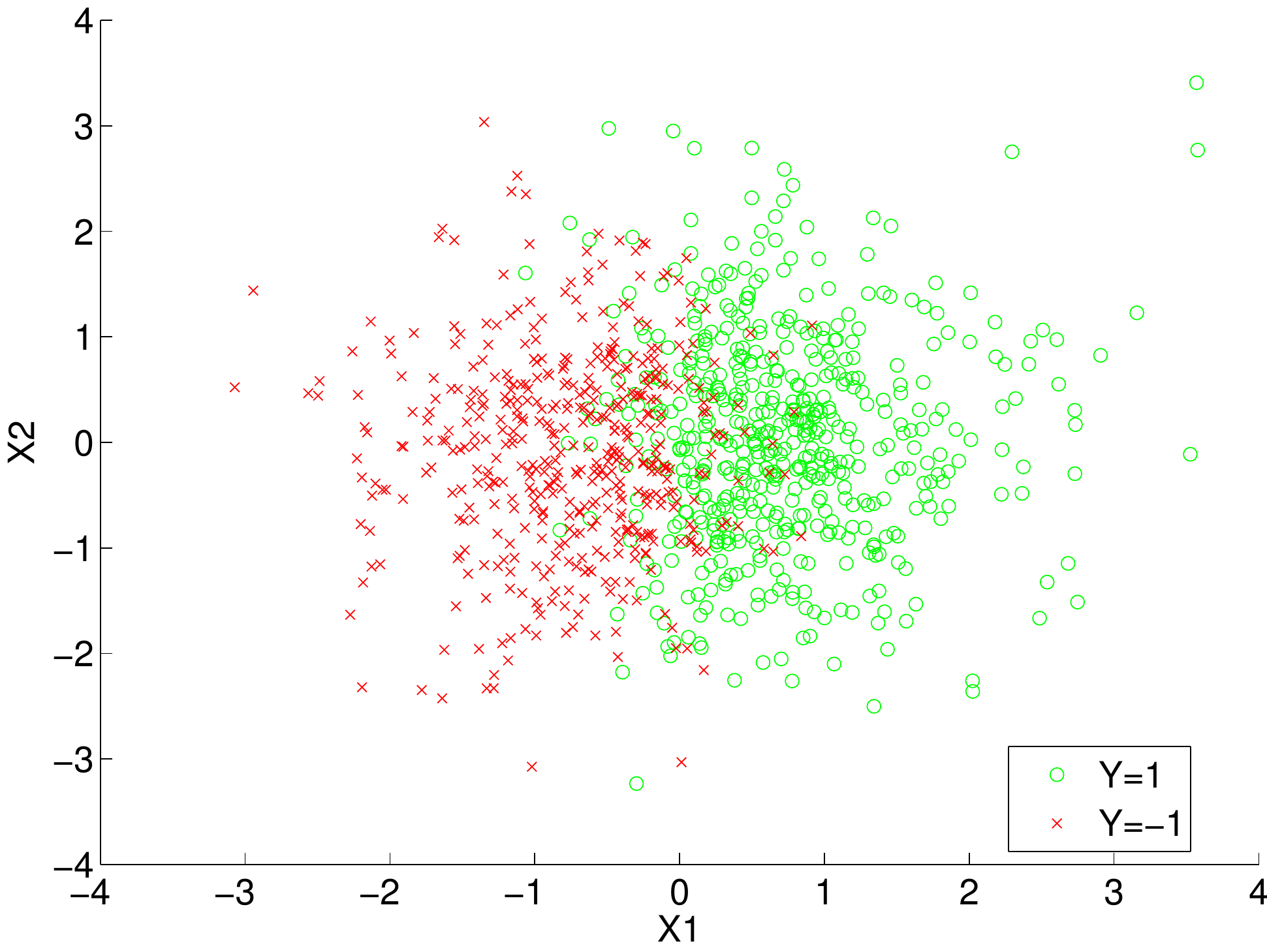}
  \caption{}
  \label{fig:sublinearity_points}
\end{subfigure}
\hfill
\begin{subfigure}[b]{0.32\textwidth}
  \includegraphics[width=\textwidth]{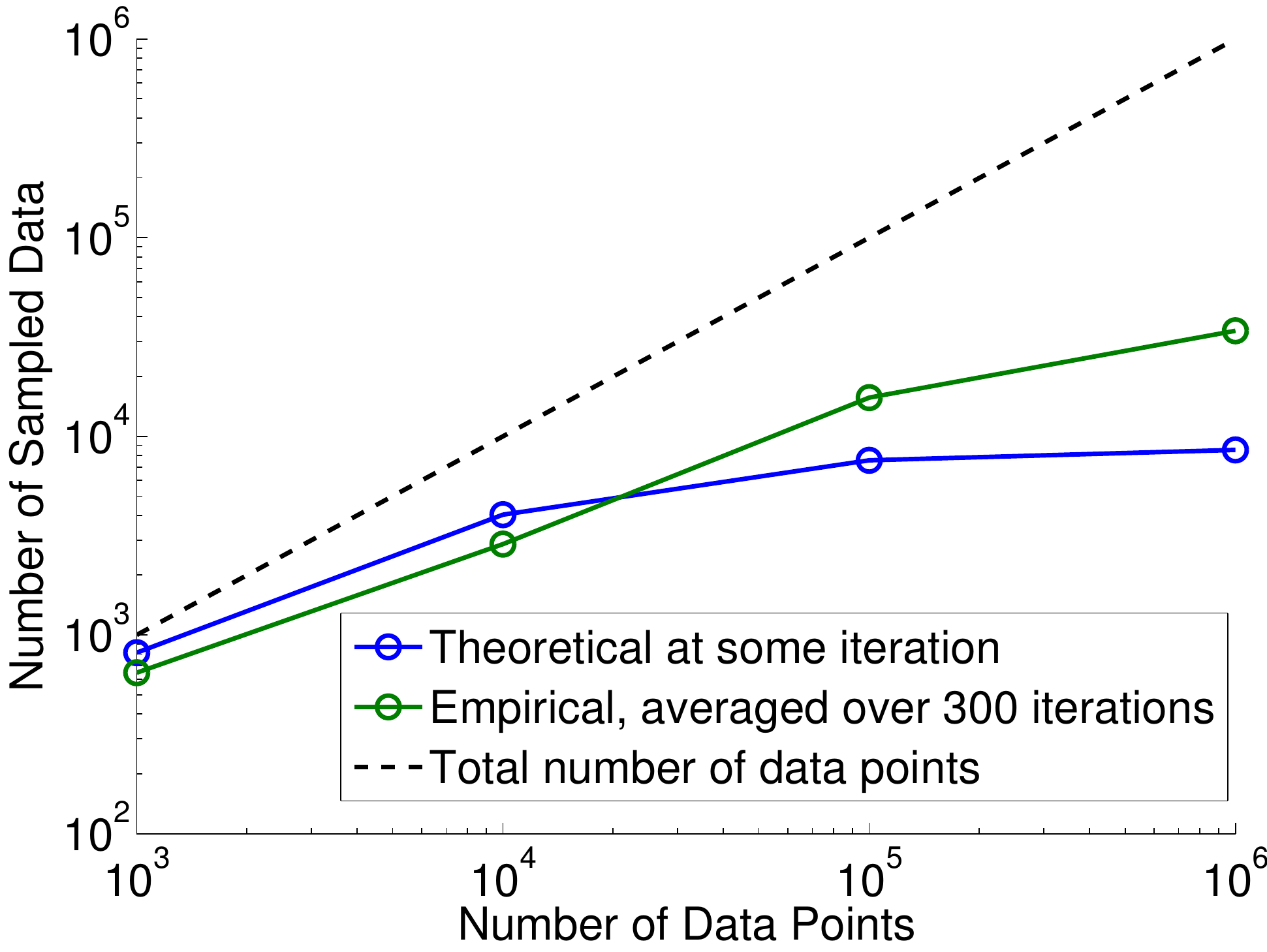}
  \caption{}
  \label{fig:sublinearity_num}
\end{subfigure}
\hfill
\begin{subfigure}[b]{0.32\textwidth}
  \includegraphics[width=\textwidth]{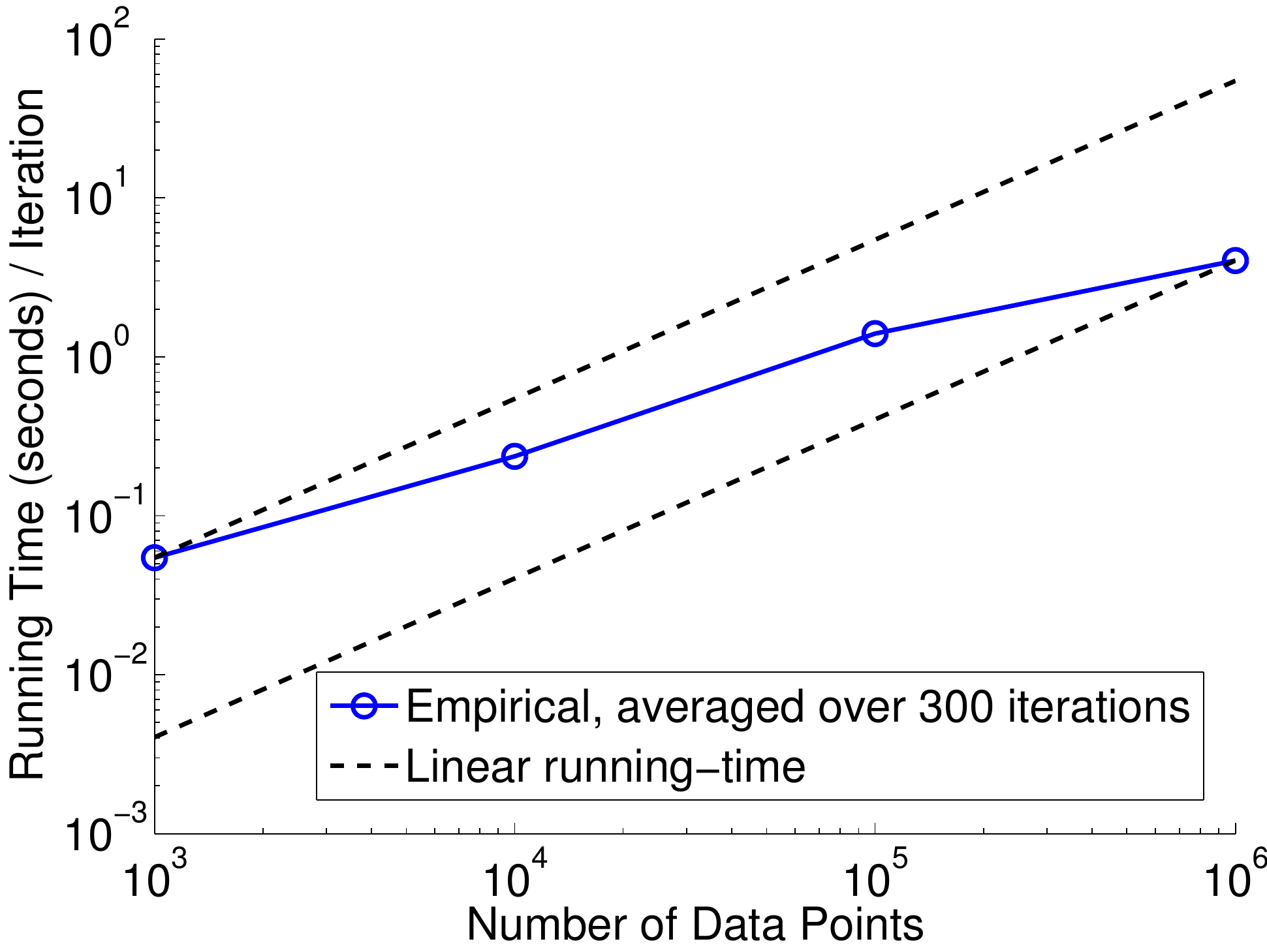}
  \caption{}
  \label{fig:sublinearity_time}
\end{subfigure}
\caption{Sublinear behavior on a synthetic data set. (a) Training data. (b) Number of subsampled data points per iteration in log-log scale. (c) Running time per iteration in log-log scale.}
\vspace{-0.3cm}
\end{figure*}

Here we empirically study the asymptotic runtime of one MH iteration with the proposed algorithm for this model. We run the same algorithm on a synthetic dataset with two input features and one binary label so that we can easily control the amount of labeled data. The dataset is illustrated in Fig.~\ref{fig:sublinearity_points}. We fix $\epsilon=0.01$, a mini-batch of 100, and a standard deviation of the proposal distribution of 0.1, and vary the number of data points. The blue line in Fig.~\ref{fig:sublinearity_num} shows the expected number of subsampled data points at a particular iteration for different sizes of the dataset in the log-log scale. The expectation is computed theoretically with Eqn. 19 in \cite{korattikara2013austerity}. We use the same current and proposed parameter value for all dataset sizes in this plot. The green line shows the empirical average number of subsampled data points over 300 iterations. Fig.~\ref{fig:sublinearity_time} shows the average running time per iteration with dashed lines indicating the behavior of a linear algorithm. Both the number of subsampled data points and running time confirm that our proposed subsample MH exhibits sublinear-linear asymptotic scaling for \textit{global} parameters given a fixed proposal distribution.

\begin{figure*}[tb]
\centering
\begin{subfigure}[b]{0.35\textwidth}
  \includegraphics[width=\textwidth]{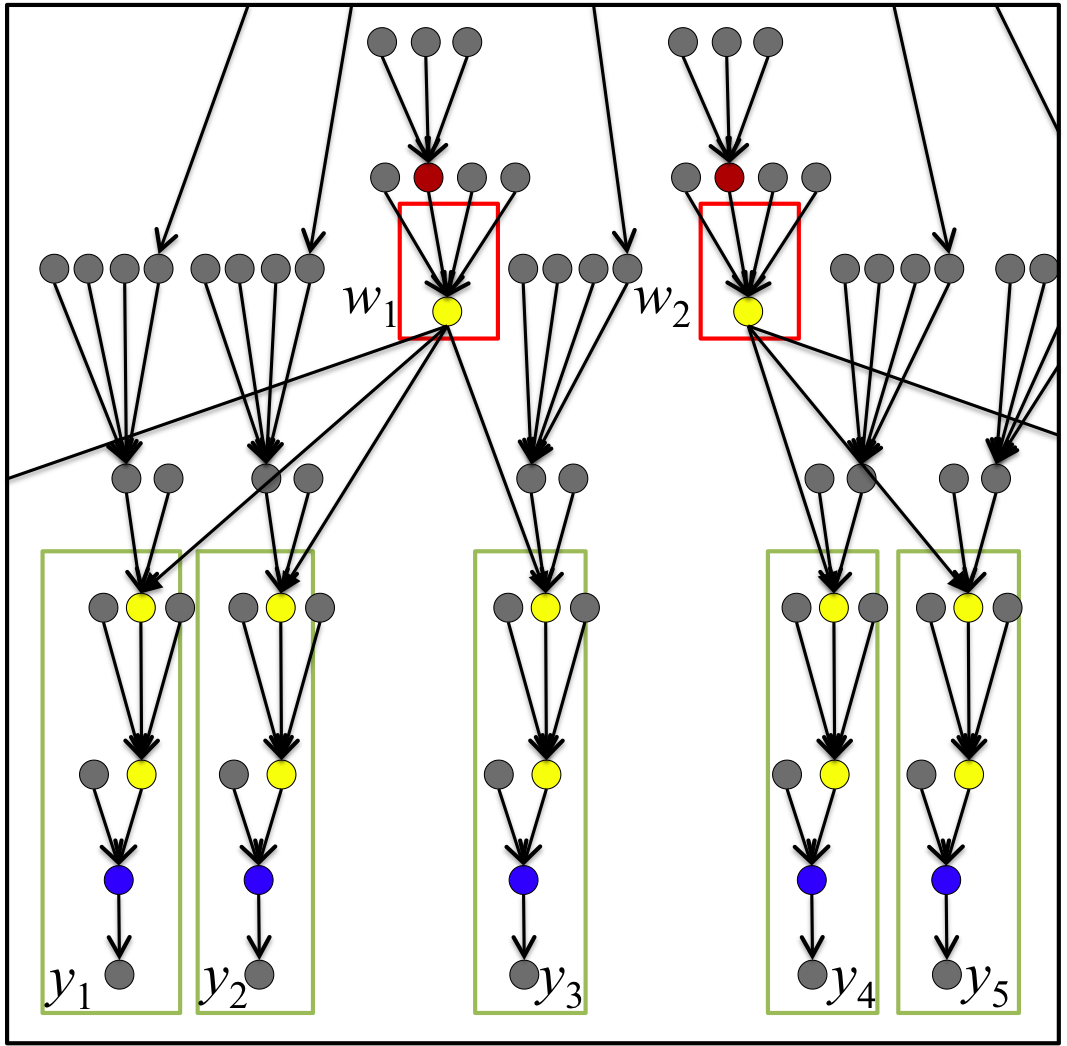}
  \caption{}
  \label{fig:scaffold_jointdplr}
\end{subfigure}
\hfill
\begin{minipage}[b]{0.16\textwidth}
\begin{subfigure}[b]{\textwidth}
  \includegraphics[width=\textwidth]{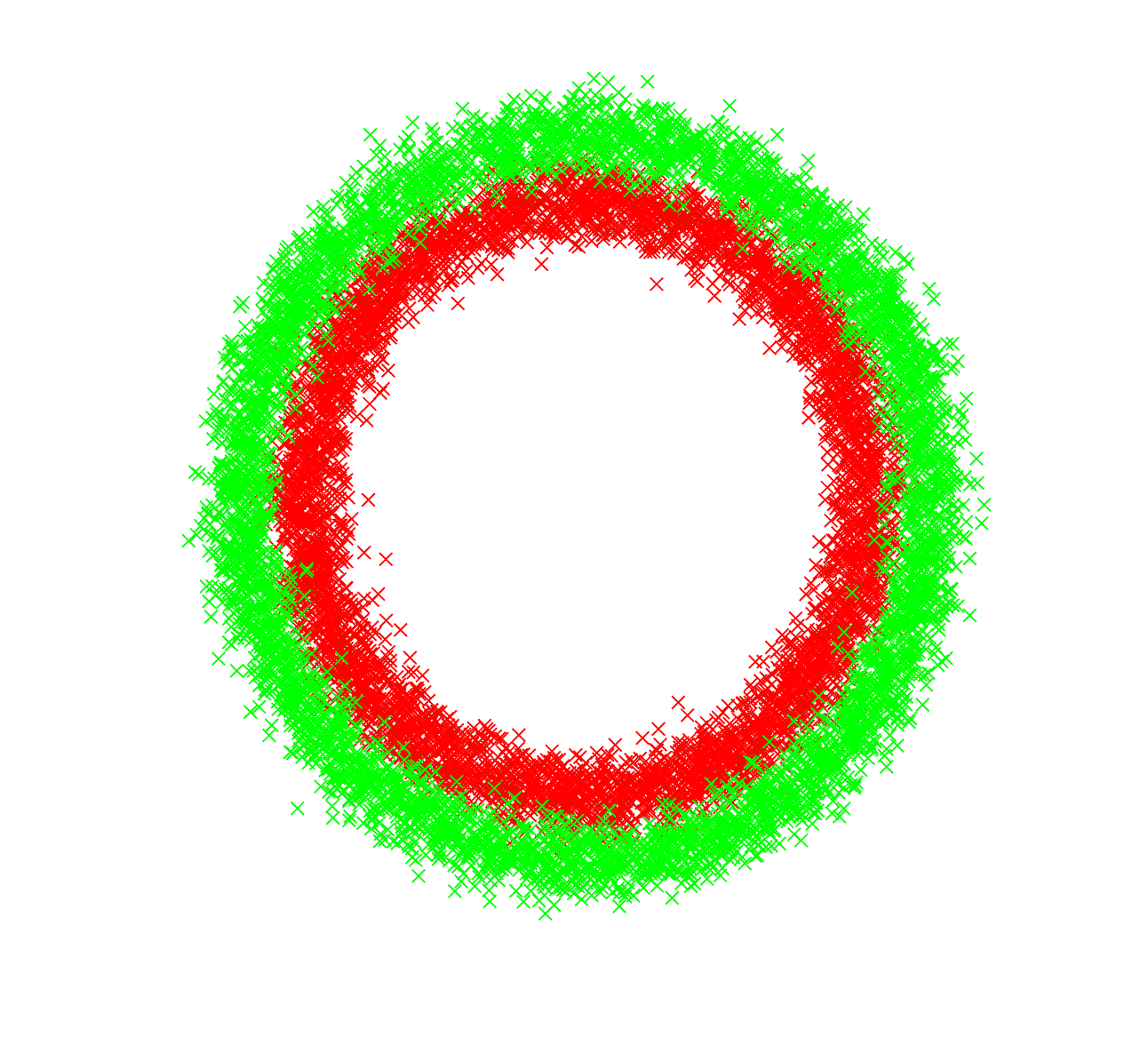}
  \caption{}
  \label{fig:jointdplr_data}
\end{subfigure}
\\
\begin{subfigure}[b]{\textwidth}
  \includegraphics[width=\textwidth]{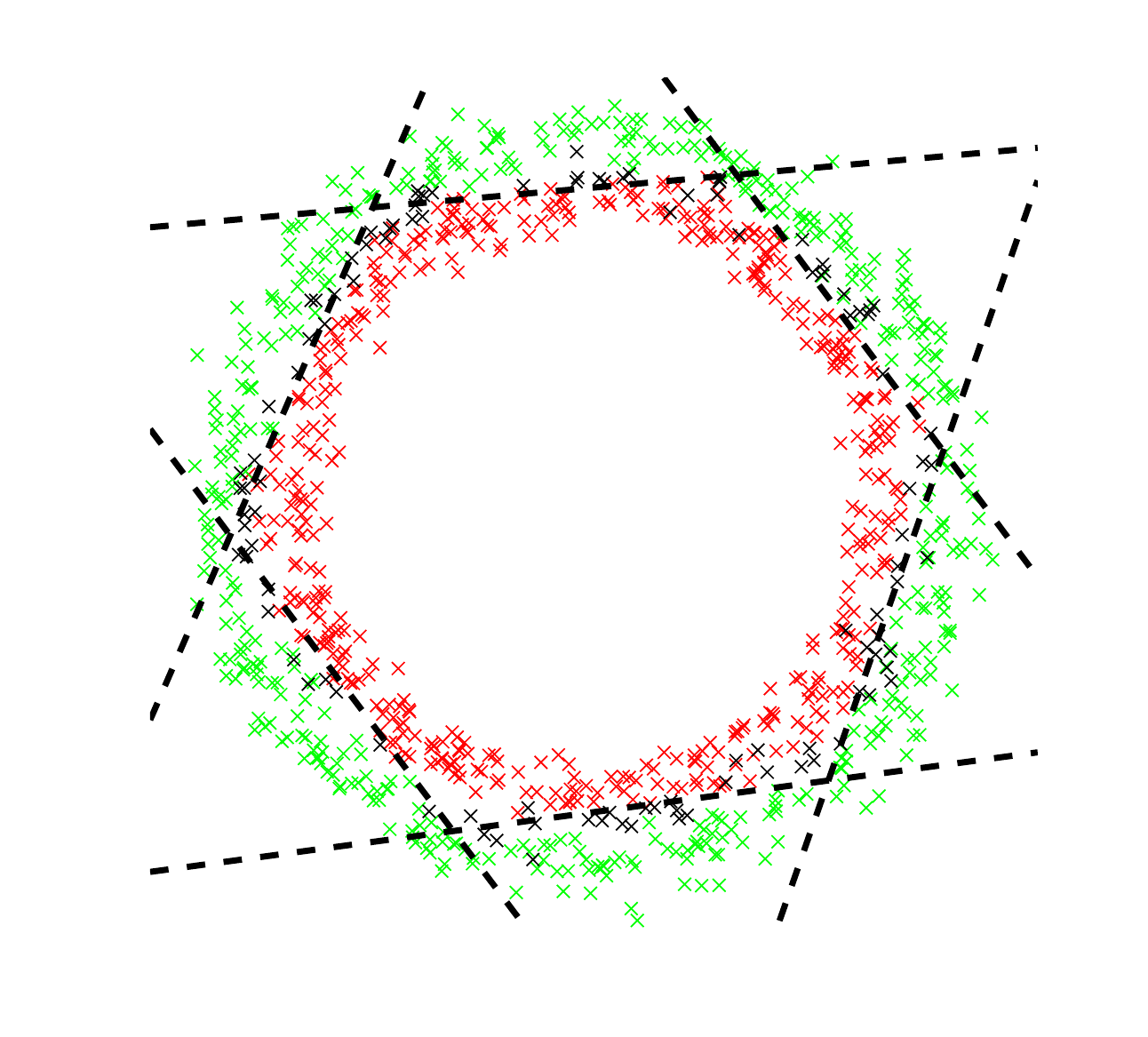}
  \caption{}
  \label{fig:jointdplr_pred}
\end{subfigure}
\end{minipage}
\hfill
\begin{subfigure}[b]{0.4\textwidth}
  \includegraphics[width=\textwidth]{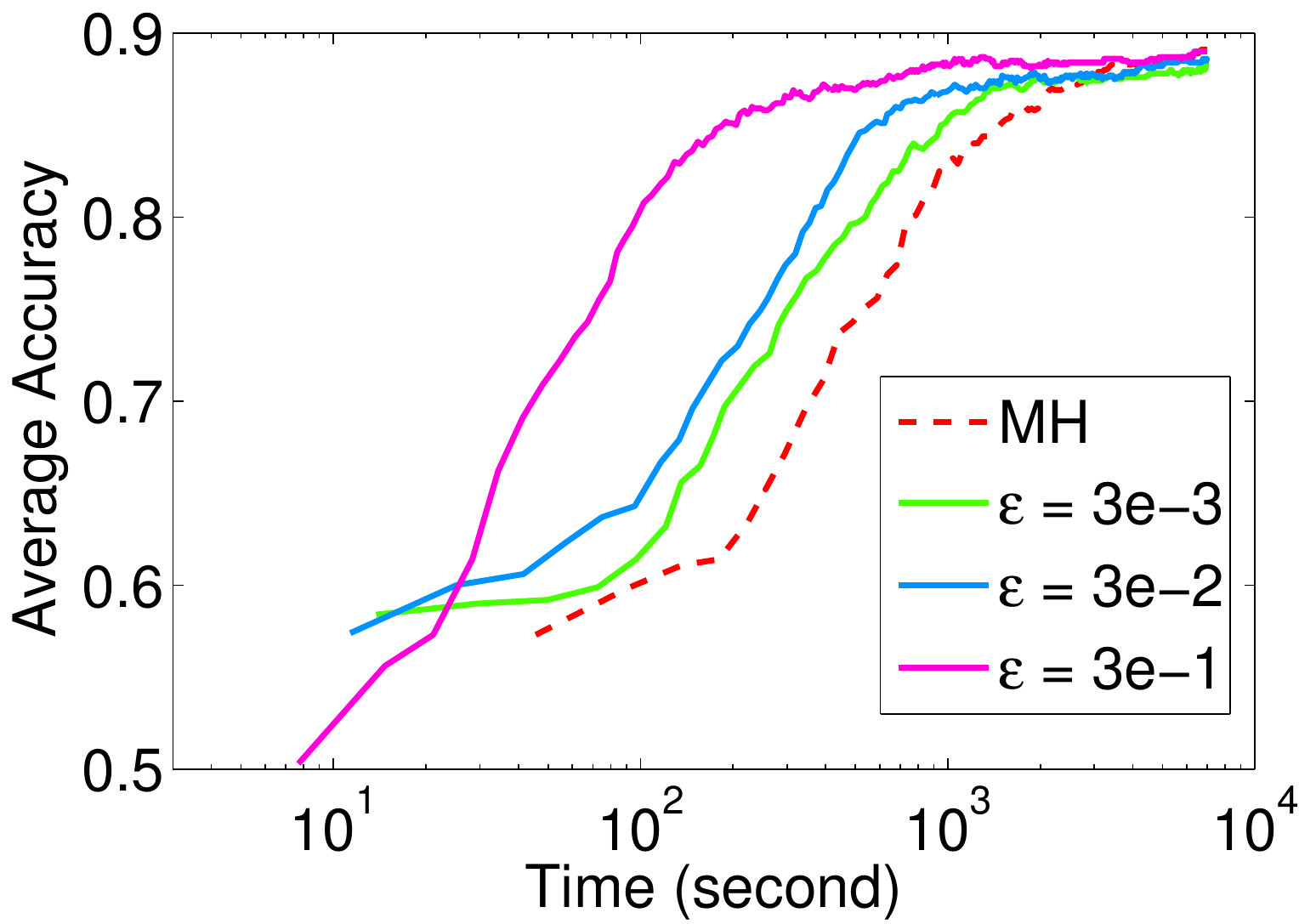}
  \caption{}
  \label{fig:jointdplr_acc}
\end{subfigure}
\caption{Joint Dirichlet Process mixture model evaluated on synthetic data. (\subref{fig:scaffold_jointdplr}) part of PET with scaffolds for two weight parameters. (\subref{fig:jointdplr_data}) training data. (\subref{fig:jointdplr_pred}) prediction on test data with $\eps=0.3$ after 2 hours. 6 clusters are found with decision boundaries (dashed lines) and misclassified points (black dots). (\subref{fig:jointdplr_acc}) Predictions accuracy vs running time in log domain.}
\label{fig:jointdplr}
\end{figure*}
\begin{figure}[tbh!]
\vspace{-0.1in}
\small\begin{Verbatim}[gobble=-30, numbers=left,numbersep=2pt, frame=single, xleftmargin=3mm,
xrightmargin = 3mm]
# fixed: dimensionality D, hypers mu_w sig_w m0 k0 v0 S0
[assume alpha (scope_include 'hypers 0 (gamma 1 1))]
[assume crp   (make_crp alpha)]
[assume z     (mem (lambda (i) (scope_include 'z i (crp))))]
[assume w (mem (lambda (z) (scope_include 'w z 
        (multivariate_normal mu_w sig_w))))]
[assume c (mem (lambda (z) 
        (make_collapsed_multivariate_normal m0 k0 v0 S0)))]
[assume x (lambda (i) ((c (z i))))]
[assume y (lambda (i x)
           (bernoulli (linear_logistic (w (z i)) x)))]
for i in 1...X: # load data
    [observe (x i) {X[i]}] #X[i] is ith feature vector
    [observe (y i) {Y[i]}] #Y[i] is ith class label

# T steps of MH for hyperparams, single-site gibbs for z,
# subsampled MH over weights for a randomly chosen expert
[infer (cycle {T} ((mh alpha all 1)
                   (gibbs z one step_z)
                   (subsampled_mh w one {Nbatch} {eps}
                                  'drift {sigma} 1)) 1)]
\end{Verbatim}
\small\begin{Verbatim}[gobble=-30, numbers=left,numbersep=2pt, frame=single, xleftmargin=3mm,
xrightmargin = 3mm]
[assume sig (scope_include 'sig 0 (sqrt (inv_gamma 5 0.05)))]
[assume phi (scope_include 'phi (beta 5 1))]
[assume h   (mem (lambda (t) (scope_include 'h t
              (if (<= t 0) 0
                  (normal (* phi (h (- t 1))) sig)))))]
[assume x (lambda (t) (normal 0 (/ (h t) 2)))]
for t in 1...T:
  [observe (x t) {X[t]}] #X[t] is the observation at time t

# state estimation via particle gibbs over subsequences of
# length L 
for n in 1...N-L:
    [infer (pgibbs h (ordered_range h h+L) P 1)]
# subsampled MH inference on the parameters
[infer (cycle ((subsampled_mh sig 0 {Nbatch} {eps} 1)
               (subsampled_mh phi 0 {Nbatch} {eps} 1)) 1)]
\end{Verbatim}
\caption{Probabilistic program containing model and inference scheme for (top) joint DPM in Sec.~\ref{sec:joint_dplr} and (bottom) stochastic volatility model in Sec.~\ref{sec:sv}.}
\label{fig:venture_jointdplr}
\vspace{-0.5cm}
\end{figure}
\begin{figure}[tbh]
\centering
\begin{subfigure}[b]{0.25\textwidth}
\includegraphics[width=\textwidth]{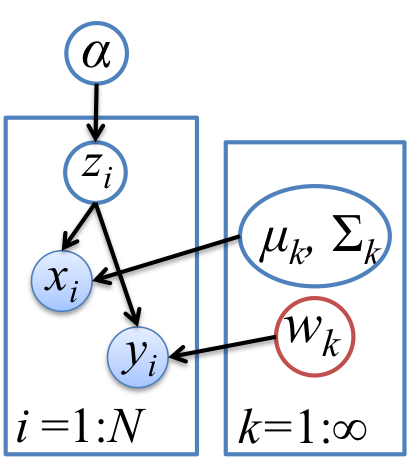}
\caption{}
\label{fig:jointdplr_plate}
\end{subfigure}
~
\begin{subfigure}[b]{0.45\textwidth}
\centering
\includegraphics[width=\textwidth]{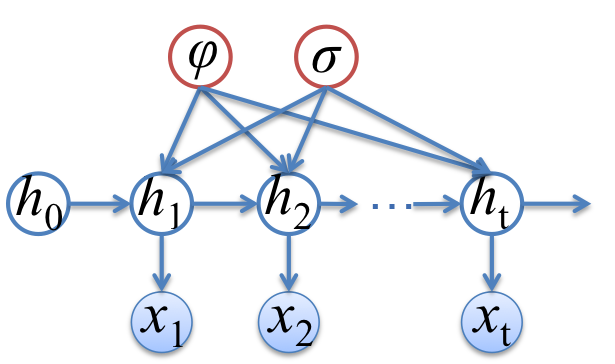}
\caption{}
\label{fig:sv_graph}
\end{subfigure}
\caption{Graphical model of (a) Joint DPM (b) stochastic volatility model. Subsampled MH applies to red nodes. The algorithm from this paper applies to problems with diverse dependency structures.}
\vspace{-0.5cm}
\end{figure}

\subsection{Joint Dirichlet Process Mixture Models}\label{sec:joint_dplr}

We also assess performance on a joint Dirichlet process mixture (JointDPM) model \citep{wade14a}, a flexible nonlinear classifier that combines logistic regressions using nonparametric Bayes. This model uses a Dirichlet process mixture of Gaussians to model input features, where each component has a distinct set of logistic regression weights to classify the vectors it contains. More formally, we have
\begin{equation}
(\bx_i, y_i)|P \overset{\textnormal{iid}}\sim f(\bx, y|P), \quad
P \sim \mathrm{DP}(\alpha P_0)
\end{equation}
where $P_0$ is the base measure. Every sample of $P$ is a countable mixture model
\begin{equation}
f(\bx, y|P) = \sum_{k=1}^{\infty} \pi_k \cN(\bx|\bmu_k, \Sig_k) \mathrm{Logit}(y|\bx, \bw_k)
\end{equation}
with each normal distribution parameter pair $(\bmu_k, \Sig_k)$ sampled from a conjugate normal-inverse-Wishart prior. The regression parameter $\bw_k$ is sampled from an isotropic Gaussian prior as in the Bayesian logistic regression model. The graphical model is shown in Fig.~\ref{fig:jointdplr_plate}. We can write the model in under 20 lines of probabilistic code (Fig.~\ref{fig:venture_jointdplr}), collapsing the component models by marginalizing out $(\mu_k,\Sig_k)$ and collapsing the DP into a CRP. A trace fragment containing two scaffolds, each containing the regression weights for one cluster, is shown in Fig.~\ref{fig:scaffold_jointdplr}. In this model, the number of simultaneous instantiations of the austerity scheme is an object of inference.

We use subsampled MH to accelerate inference over the parameters $w_k$ for each cluster's logistic regression model. The CRP hyperparameter $\alpha$ and $N$ component assignment variables $z_i$ must also be inferred, but as inference for these variables already requires constant time due to properties of the PET, approximate transitions are only used for the $w_k$s. We set the parameters of our inference program to allocate roughly equivalent computation time to sampling $w_k$ and to a series of transitions to randomly chosen $z_i$'s; the balance is struck using the \verb|step_z| parameter (see Fig.~\ref{fig:venture_jointdplr}). Because subsampled MH runs faster than MH, the $w$'s are updated more frequently than standard MH to maintain the balance.


\begin{figure*}[tbh!]
\centering
\begin{subfigure}[b]{0.3\textwidth}
  \includegraphics[width=\textwidth]{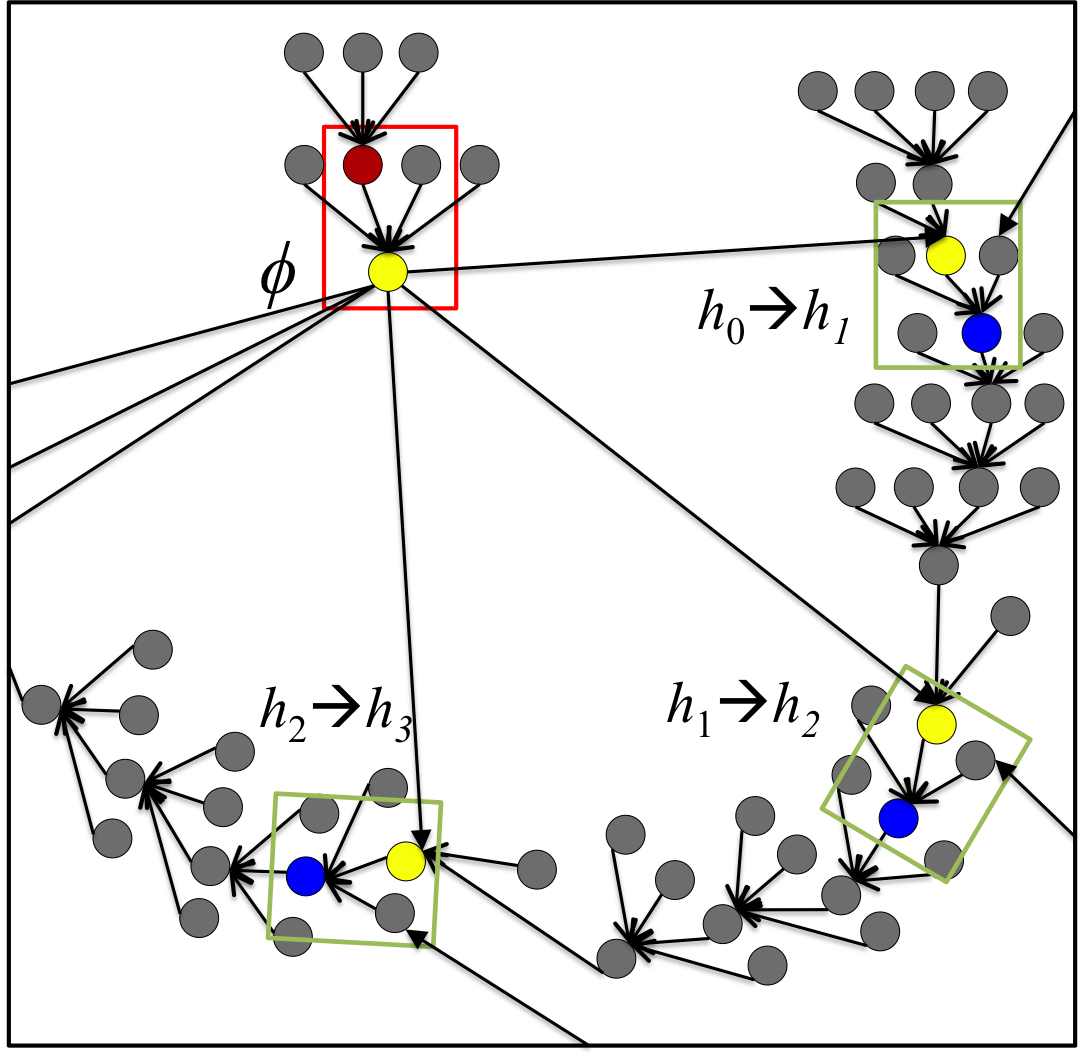}
  \caption{}
  \label{fig:scaffold_sv}
\end{subfigure}
\hfill
\begin{minipage}[b]{0.22\textwidth}
\begin{subfigure}[b]{\textwidth}
  \includegraphics[width=\textwidth]{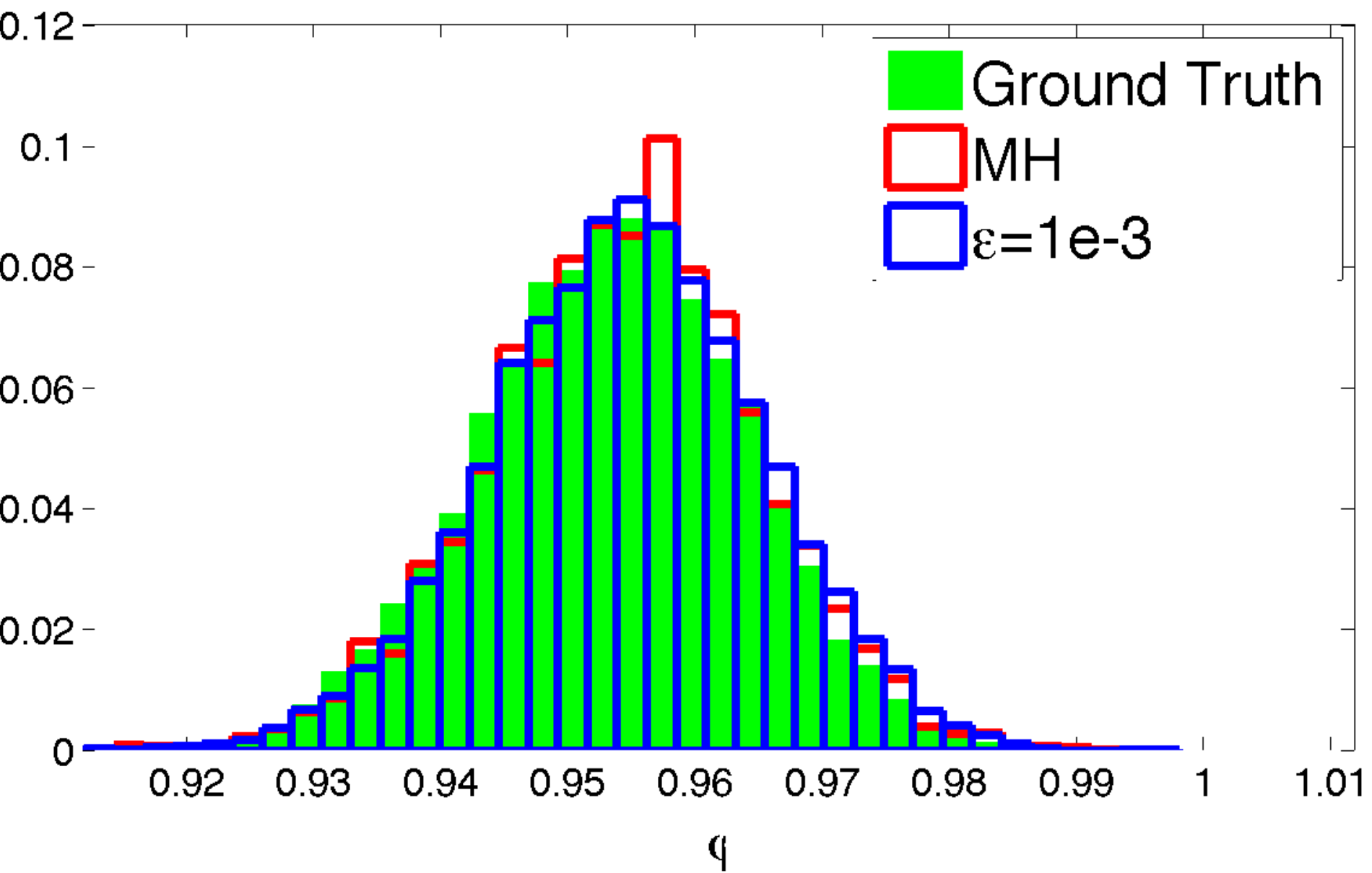}
  \caption{Histogram of $\phi$.}
  \label{fig:sv_hist_phi}
\end{subfigure}
\\
\begin{subfigure}[b]{\textwidth}
  \includegraphics[width=\textwidth]{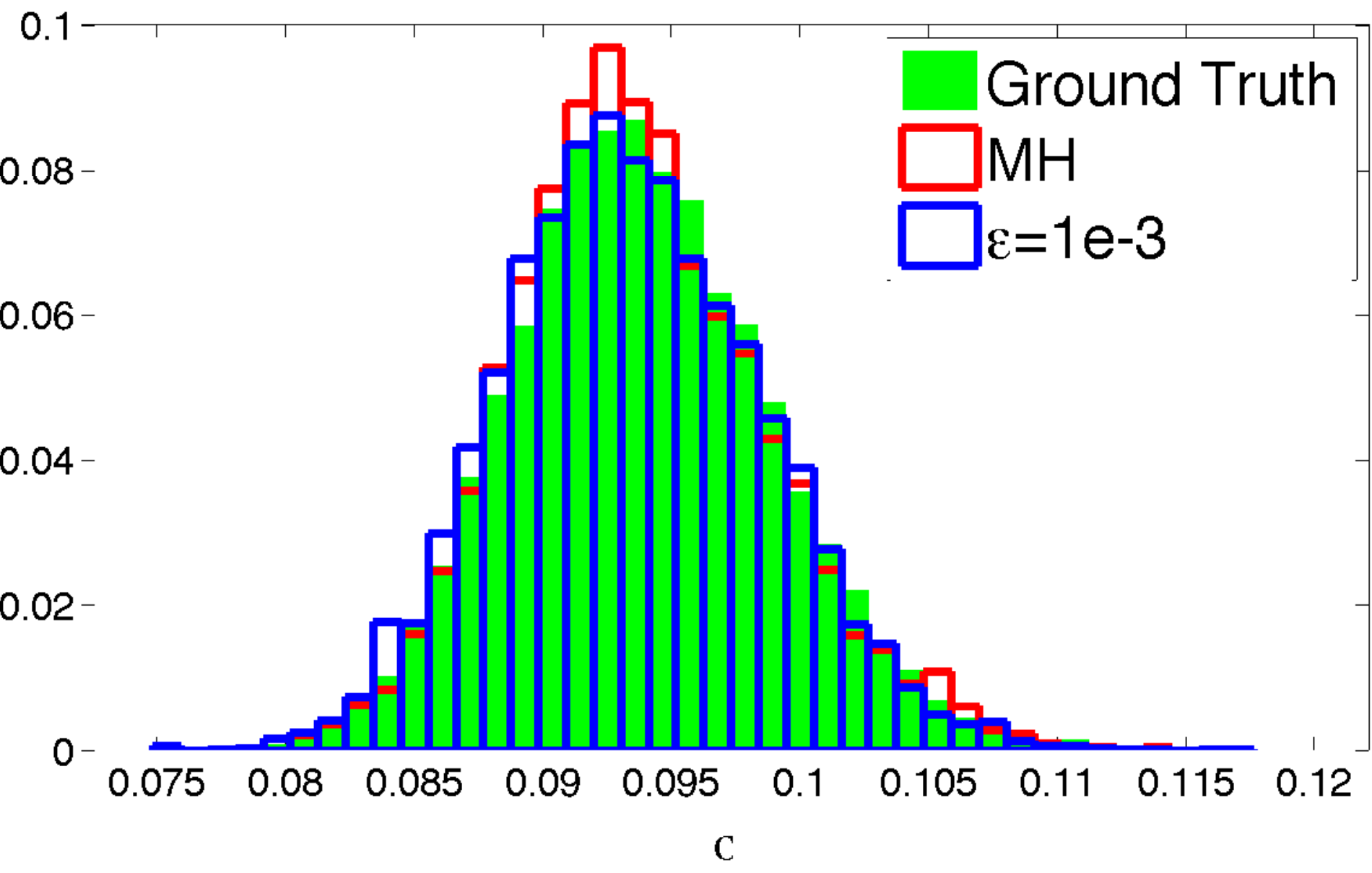}
  \caption{Histogram of $\sg$.}
  \label{fig:sv_hist_sigma}
\end{subfigure}
\end{minipage}
\hfill
\begin{subfigure}[b]{0.4\textwidth}
  \includegraphics[width=\textwidth]{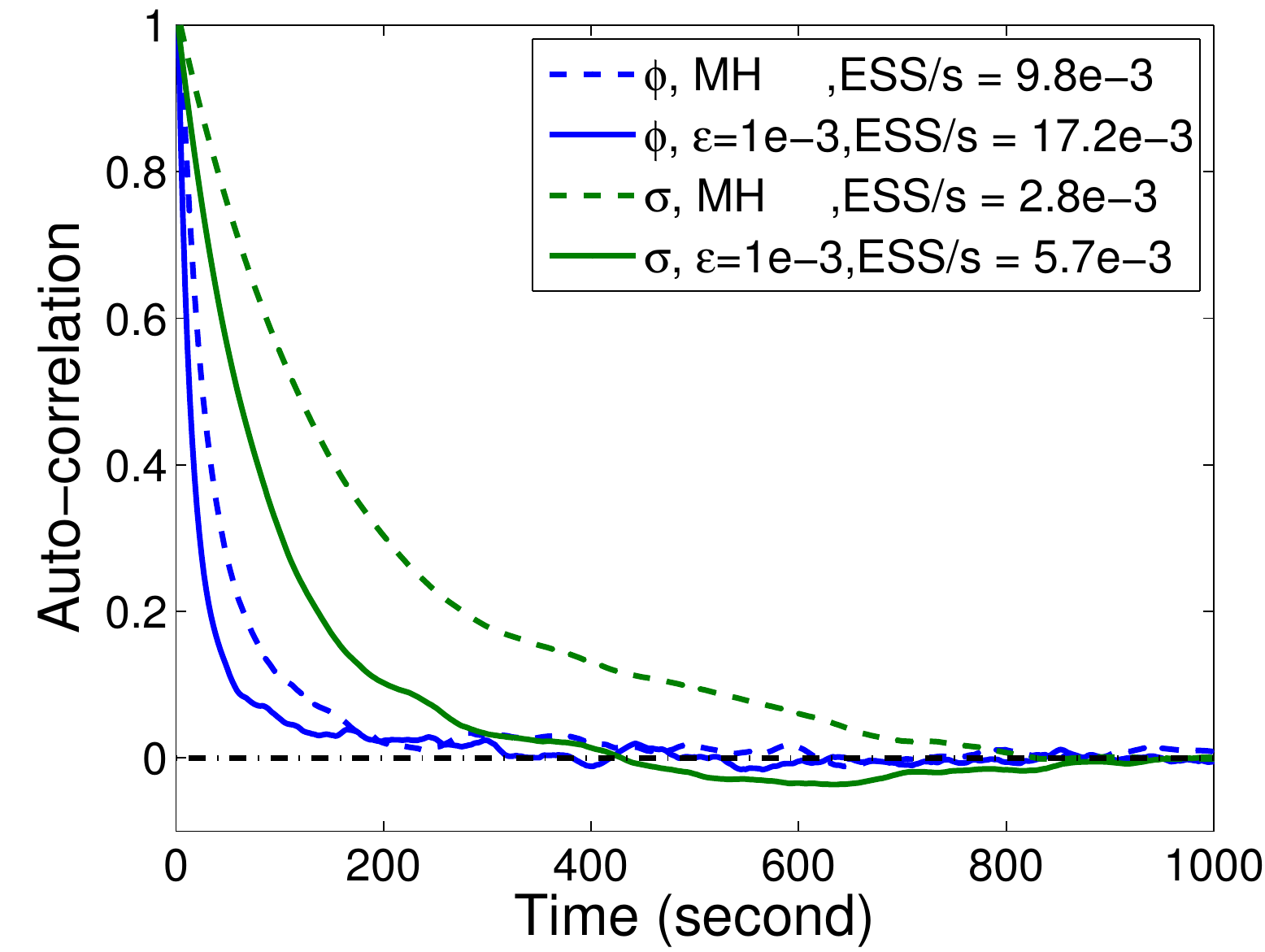}
  \caption{}
  \label{fig:sv_autocorr}
\end{subfigure}
\caption{Stochastic Volatility Model. (\subref{fig:scaffold_sv}): part of PET and the scaffold for sampling $\phi$. (\subref{fig:sv_hist_phi}, \subref{fig:sv_hist_sigma}): Histogram of the samples of $\phi$ and $\sg$ from ground truth (green), exact MH (red) and subsampled MH with $\eps$=1e-3 (blue). (\subref{fig:sv_autocorr}): Auto-correlation of samples for exact (dashed) and subsampled MH (solid). ESS per second shown in the legend.}
\label{fig:sv}
\vspace{-0.4cm}
\end{figure*}

We apply JointDPM to a synthetic data set with 10,000 data points as shown in Fig.~\ref{fig:jointdplr_data}.
A snapshot of the prediction results on another 1,000 test points and the accuracy of the prediction averaged over the Markov chain are shown in Fig.~\ref{fig:jointdplr_pred} and \ref{fig:jointdplr_acc}. The sublinear algorithm with $\eps=0.3$ reaches the same accuracy as exact MH in 10x less time.

\subsection{Joint Parameter and State Estimation in a State-Space Model}\label{sec:sv}
We also applied our implementation to a stochastic volatility model for a time series with two parameters:
\begin{equation*}
x_t = \exp(h_t/2) \eps_t,~h_t \sim \cN(\phi h_{t-1}, \sg^2),~\eps_t \overset{\textnormal{iid}}\sim \cN(0, 1)
\end{equation*}
where we set $h_0=0$ and assign a Beta(5,1) and an inverse Gamma(5, 0.05) distribution for parameter $\phi$ and $\sg^2$ respectively. This model is a state-space model and has unknown hidden states as well as unknown parameters. Fig.~\ref{fig:sv_graph} shows the graphical model. Fig.~\ref{fig:scaffold_sv} illustrates the PET of the model with $t=1,2,3$. Note that there are dependencies between the subsampled local partitions in this problem, transitions from $h_{t-1}$ to $h_t$. The dynamics are sensitive to the precise values of $\phi$ and $\sg^2$ and also the hidden states, yielding a challenging joint inference problem.

We generate a synthetic data set of 200 series of length 5 with correlation $\phi=0.95$ and noise $\sg=0.1$. We apply particle Gibbs to sample latent states and (subsampled) MH to sample parameters $\phi$ and $\sg$. Fig.~\ref{fig:sv_hist_phi} and \subref{fig:sv_hist_sigma} show the histogram of the samples after the burn-in period, and Fig.~\ref{fig:sv_autocorr} shows the autocorrelation of the samples measured in running time. Due to the high correlation between latent state variables, the overall mixing rate highly depends on the mixing rate in $h_t$'s. Therefore we assign 10 more times of computation time to sample $h_t$ than other variables. We observe that the gain of subsampled MH is not as significant as previous experiments due to the slow mixing of latent states. Nevertheless, subsampled MH still obtains about twice the efficiency of exact MH without introducing significant bias.

\section{Discussion}
\vspace{-0.1in} This paper shows that it is feasible to define sublinear approximate MH transitions for highly-coupled variables in general probabilistic programs and integrate them into an inference programming language. The results also suggest that it is useful: a single unoptimized implementation applies to a broad class of problems, going beyond previous approximate MH schemes, and yields significant (2x-10x) improvements in runtime at no cost in accuracy.

The algorithm presented here uses frequentist statistical inference to accelerate Bayesian inference. It is interesting to consider alternatives to Algorithm 1, where the sequential test is replaced with a model-based inference --- potentially itself written as a probabilistic program. It may also be interesting to consider probabilistic programming adaptations of other techniques for accelerating inference stochastically ignoring or suppressing dependencies, such as decayed MCMC filtering \citep{golightly2014delayed}.

Probabilistic programs can represent model classes, datasets, queries and custom inference strategies from many different application domains. Some domains will require faithful, fully Bayesian inference. In these cases, approximate MH may be useful for accelerating burn-in. For other problems, such as parameter estimation for motion models in robotics or machine learning from internet user behavior, speed may be preferred to accuracy, and results from sublinear approximate MH transitions may be adequate on their own. We hope that by integrating sublinear, approximate transitions into a higher-order probabilistic programming system, we have taken a step towards making it more practical to use probabilistic programming in both these kinds of applications.

\bibliographystyle{plainnat}
\bibliography{Refs_yutian} %





\appendix

\section{Proof of Theorems \ref{thm:max_error}}
\begin{proof}
Given the value of $(\ta, \ta^*)$, if all numbers in the set $\{l_i\}$ all equal, Alg.~\ref{alg:submh_mu} would always produce the correct decision and therefore $P_a(\ta, \ta^*)=P_{a,\eps}(\ta, \ta^*)$. We only consider the case when all the numbers are not equal from now on. Since the size of the set $\{l_i\}$ is finite, there exists a positive minimal value for any positive estimated standard deviation obtained from a subset of $\{l_i\}$:
\begin{equation}
s_l^*(\ta,\ta^*) \defeq \min_{S\subseteq(\{l_i\}): s_l(S)>0} s_l(S) > 0.\nn
\end{equation}

In the accept/reject phase of a MH iteration, given the uniform random variable $u$, the sequential test algorithm will make a wrong decision if at some step in the sequence of tests $\hat{\mu} - \mu_0 > s t_{n-1}^{-1}(1-\eps)$ while the true mean $\mu \leq \mu_0$ or $\hat{\mu} - \mu_0 < -s t_{n-1}^{-1}(1-\eps)$ while $\mu > \mu_0$, where $t_{n}^{-1}$ is the inverse CDF of student t distribution with a degree of freedom $n$. Letting $C(\ta, \ta^*)\defeq\max_{i}\{|l_i|\}$, as we do a test only when $s_l>0$, we can bound the probability of making an error at step $t$ by applying the one side Hoeffding's inequality with adjustment for sampling without replacement \citep{serfling1974probability} as
\begin{align}
& P\left(\hat{\mu} - \mu_0 > s t_{n-1}^{-1}(1-\eps)\right) \nn\\
& \leq P\left(\hat{\mu} - \mu > \frac{s_l^*}{\sqrt{n}}\sqrt{1-\frac{n-1}{N-1}} t_{n-1}^{-1}(1-\eps)\right) \nn\\
& \leq \exp\left(-\ha \left(\frac{s_l^* t_{n-1}^{-1}(1-\eps)}{C}\right)^2 \right) \nn\\
& \defeq f_1(\eps, n, \ta, \ta^*)
\end{align}
when $\mu \leq \mu_0$. In the fist inequality we plugged in the assumption $\mu \leq \mu_0$, the definition of $s$ in Step 8 of Alg.~\ref{alg:submh_mu} and the lower bound of $s_l$. In the second inequality we applied the concentration bound. It is obvious to observe that the error is also upper bounded by $f_1$ when $\mu > \mu_0$. Notice that $f_1(\eps,n,\ta,\ta^*) \rightarrow 0$ as $\eps \rightarrow 0, \forall n, \ta, \ta^*$.

We can furthur bound the total probability of making a wrong decision in the sequential test given $u$, denoted as $\cE(u,\eps)$, using the union bound:
\begin{align}
\cE(u, \eps, \ta, \ta^*) &= P\left(\cup_t \{\text{wrong decision at step } t\}\right) \nn\\
& \leq \sum_{t=1}^{\lceil N/n \rceil} f_1(\eps, mt, \ta, \ta^*) \defeq f_2(\eps, \ta, \ta^*)
\end{align}
We omit the dependency of $f_2$ on $N$ and $m$. Again, we have $f_2(\eps,\ta,\ta^*) \rightarrow 0$ as $\eps \rightarrow 0, \forall \ta, \ta^*$.

Denote the acceptance probability of our approximate Markov chain at $(\ta, \ta^*)$ as $P_{a,\eps}$ and we can bound its error as
\begin{align}
& |P_{a,\eps}(\ta, \ta^*) - P_a(\ta, \ta^*)| \nn\\
&= \left| \int_0^1 P(A|u) \D u - \int_0^{P_a} \D u \right| \nn\\
&= \left| \int_{P_a}^1 P(A|u) \D u -\int_0^{P_a} (1-P(A|u))\D u \right| \nn\\
&\leq \int_0^1 \cE(u, \eps, \ta, \ta^*) \D u \leq f_2(\eps, \ta, \ta^*)\label{eq:error_P_a}
\end{align}
where A denotes the event that the sequential test procedure accepts the proposal. So $P(A|u)$ with $u>P_a$ is the probability of accepting a proposal while we should reject it and $1-P(A|u)$ with $u<P_a$ is the probability of rejecting a proposal when we should accept it.

Now let us consider the two conditions in Theorem \ref{thm:max_error} separately.

Under condition (1), the value of $l_i$ as a function of $(\ta, \ta^*)$ is continuous. Therefore, the functions $s^*$ and $C$ are also continuous w.r.t. $(\ta, \ta^*)$. So is the upper bound $f_2(\ta, \ta^*)$.

Combined with the condition that $\Theta$ is compact, we can conclude the proof by claiming that function $f_2$ will achieve its maximum in the domain of $\Theta \times \Theta$, denoted as $\de(\eps)$, and because the function $f_2$ approaches 0 everywhere in the compact set $\Theta \times \Theta$ as $\eps\rightarrow 0$, $\de(\eps)$ also approaches 0 with $\eps$.

Under condition (2), since the domain of $\ta$ is finite, there exists a maximum value of $f_2(\eps,\ta,\ta^*)$ over $(\ta,\ta^*)$ for any $\eps$. Let
\begin{equation}
\delta(\eps) \defeq \max_{\ta,\ta^*\in\Theta\times\Theta} f_2(\eps, \ta, \ta^*)
\end{equation}
Since $\forall \ta,\ta^*\in\Theta\times\Theta$, $f_2(\eps, \ta, \ta^*)\rightarrow 0$ as $\eps\rightarrow 0$, it follows that $\delta(\eps)\rightarrow 0$ as $\eps\rightarrow 0$.
\end{proof}

\section{Proof of Corollary \ref{cor:ergodicity}}
\begin{proof}
Denote the proposal distribution with $q(\ta, \ta')$. For a variable defined in a compact space $\Theta$ with measure $\Omega$, the error of expected rejection probability of the approximate Markov chain is bounded by
\begin{equation}
|P_{r,\eps} - P_r| = \left| \int_{\ta'} (-P_{a,\eps} + P_a) q(\ta, \ta') \D\Omega(\ta') \right| \leq \de(\eps)
\end{equation}
The transition kernel of M-H is $\cT(\ta, \ta') = P_a q(\ta, \ta') + P_r \de(\ta, \ta')$, where $\de(\ta, \ta')$ denotes the Dirac delta function. We can bound the total variation distance between the approximate M-H kernel and the exact kernel as
\begin{align}
& \|\cT_{\eps}(\ta,\cdot) - \cT(\ta,\cdot)\|_{\mathrm{TV}} \nn\\
&=\ha \int_{\ta'} \Big| \left(P_{a,\eps}(\ta, \ta^*) - P_a(\ta, \ta')\right)q(\ta, \ta') \nn\\
&\quad\quad\quad\quad + (P_{r,\eps} - P_r)\de(\ta, \ta')\Big| \D\Omega(\ta') \nn\\
&\leq \de(\eps), \forall \ta \in \Theta
\end{align}
Since $\de(\eps)\rightarrow 0$ as $\eps\rightarrow 0$, for any sufficiently small $\eps$ we can apply the Lemma 3.6 of \cite{pillai2014ergodicity} to prove the uniform ergodicity and obtain the convergence rate.
\end{proof}

\end{document}